\documentclass[11pt]{article}

\usepackage{packages_macros_commands}

\begin{document}
\title{Learning Rates for Kernel-Based Expectile Regression}

\author{Muhammad Farooq and Ingo Steinwart\\
Institute for Stochastics and Applications\\
Faculty 8: Mathematics and Physics\\
University of Stuttgart\\
D-70569 Stuttgart Germany \\
\texttt{\small \{muhammad.farooq\footnote{This research is supported by Higher Education Commission (HEC) Pakistan 
	(PS/OS-I/Batch- 2012/Germany/2012/3449) and German Academic Exchange Service (DAAD) scholarship program/-ID50015451.},
	ingo.steinwart\}@mathematik.uni-stuttgart.de}
}

\maketitle

\begin{abstract}\label{sec-head-abstract}
Conditional expectiles are becoming an increasingly important tool in finance as well as 
in other areas of applications. We analyse a support vector machine type approach for 
estimating conditional expectiles and establish learning rates that are minimax optimal 
modulo a logarithmic factor if Gaussian RBF kernels are used and the desired expectile
is smooth in a Besov sense. As a special case, our learning rates improve the best known 
rates for kernel-based least squares regression in this scenario. Key ingredients of our
statistical analysis are a general calibration inequality for the asymmetric least squares
loss, a corresponding variance bound as well as an improved entropy number bound for 
Gaussian RBF kernels.
\end{abstract}

\section{Introduction}\label{sec-head-intro}
Given i.i.d samples $D:=((x_1,y_1), \ldots, (x_n, y_n))$ drawn from some unknown probability
distribution $\mathrm{P}$ on $X \times Y$, where $X$ is an arbitrary set and 
$Y \subset \mathbb{R}$,  the goal to explore the conditional distribution of $Y$ given
$x \in X$ beyond the center of the distribution can be achieved by using both quantile 
and expectile regression. The well-known quantiles are obtained by minimizing asymmetric 
least absolute deviation (ALAD) loss function proposed by \cite{koenker1978regression}, 
whereas  expectiles are computed by minimizing asymmetric least square (ALS) loss function 
\begin{equation}\label{lossALS}
\begin{aligned}
 L_{\tau}(y,t)=\left\{\begin{array}{ll}
	(1-\tau) (y-t)^2\,, & \hspace*{6ex} \text{if} \hspace*{2ex} y < t\,,\\ 
	\tau (y-t)^2\,,     & \hspace*{6ex} \text{if} \hspace*{2ex} y \geqslant t\,,
\end{array}\right.
\end{aligned}
\end{equation}
for all $t \in \mathbb{R}$ and a fixed $\tau \in (0,1)$, see primarily \cite{newey1987asymmetric}
and also \cite{efron1991regression, abdous1995relating} for further references.
These expectiles have attracted considerable attention in recent years and have been applied 
successfully in many areas, for instance, in demography \cite{schnabel2009analysis}, in education
\cite{sobotka2013estimating} and extensively in finance 
\cite{wang2011measuring,hamidi2014dynamic,xu2016nonparametric, kim2016nonlinear}. In fact, it has
recently been shown (see, e.g.~\cite{BeKlMuGi14a}, \cite{StPaWiZh14a}) that expectiles are the 
only risk measures that enjoy the properties of coherence and  elicitability, 
see \cite{gneiting2011making}, and therefore they have been suggested as potentially  better
alternative  to both Value at Risk (VaR)  and Expected Shortfall (ES), see e.g.~\cite{taylor2008estimating, ziegel2014coherence, BeKlMuGi14a}.  In order to see more applications of 
expectiles, we refer the interested readers to, e.g.~\cite{aragon2005conditional, stahlschmidt2014expectile, guler2014mincer}.

Both quantiles and expectiles are special cases of so-called  asymmetric $M$-estimators
(see \cite{breckling1988m}) and there exists one-to-one mapping between them 
(see, e.g.~\cite{efron1991regression}, \cite{abdous1995relating} and \cite{yao1996asymmetric}
),
in general, however, expectiles do not coincide with quantiles. Hence, the choice between expectiles 
and quantiles mainly depends on the application at hand, as it is the case in the duality 
between the mean and the median. For example, if the goal is  to estimate the (conditional) threshold for 
which $\tau$-fraction of (conditional) observations lie below that threshold, then $\tau$-quantile regression 
is the right choice. On the other hand, if one is interested to estimate the (conditional) threshold 
for which the average of below threshold excess information (deviations of observations from threshold) 
is $k$ times larger then above that threshold, then the 
$\tau$-expectile regression is a preferable choice with $\tau=\frac{k}{k-1}$, see \cite[p. 823]{newey1987asymmetric}. 
In other words, 
the focus in quantiles is the ordering of observations while expectiles account magnitude of the observations,
which makes expectiles sensitive to the extreme values of the distribution and this sensitivity thus play 
a key role in computing the ES in finance. Since, estimating  
expectiles is computationally more efficient than quantiles, one can however use expectiles as a promising
surrogate of quantiles in the situation where one is only interested to explore 
the conditional distribution.

As already mentioned above, $\t$-expectiles can be computed with the help of asymmetric 
risks 
%
\begin{align}\label{sec1-eq-risk}
 \frisk{f}:=\int_{X\times Y} L_{\tau}(y, f(x)) dP(x,y)=\int_X \int_Y  L_\t (y,f(x))dP(y|x)dP_X(x)\,,
\end{align}
where $P$ is the data generating distribution on $X\times Y$ and $f:X\to \R$ is some 
predictor. To be more precise,  there exists a $P_X$-almost surely unique function
$f_{L_{\tau},P}^{\star}$ satisfying 
%
%
\begin{align*}
 \frisk{f_{L_{\tau},P}^{\star}}= \orisk:=\inf \{\frisk{f} \, |\, f: X \to \mbb{R} \mbox{ measurable} \}\,,
\end{align*}
and $f_{L_{\tau},P}^{\star}(x)$ equals 
$\tau$-expectile of the conditional distribution
$P(\cdot|x)$ for $P_X$-almost all $x\in X$.

%
%

Some semiparametric and nonparametric methods for  estimating conditional $\tau$-expectiles
with the help of 
empirical $L_{\tau}$-risk have already been proposed 
in literature,
see e.g.~\cite{ sobotka2012geoadditive, yao1996asymmetric, yang2014nonparametric}
for further details. Recently,  \cite{farooq2015svm} proposed an another nonparametric estimation method
that belongs to the family of so-called kernel based regularized empirical risk minimization,
which solves an optimization problem of the form
\begin{equation}\label{SVM-expectile}
 f_{D, \lb} = \arg \underset{f\in H}{\min} \, \lb \snorm{f}_{H}^2 + \mathcal{R}_{L_{\tau},D}(f)\,.
\end{equation}
Here, $\lb > 0$ is a user specified regularization parameter, $H$ is a reproducing kernel Hilbert space (RKHS) over $X$
with reproducing kernel $k$ (see, e.g.~\cite{aronszajn1950theory} and \cite[Chapter 4.2]{steinwart2008support}) and $\mathcal{R}_{L_{\tau},D}(f)$ denotes 
the empirical risk of $f$, that is
\begin{equation*}
 \mathcal{R}_{L_{\tau},D}(f)=\frac{1}{n}\sum_{i=1}^n L_{\tau}(y_i, f(x_i))\,.
\end{equation*}
Since the ALS loss $L_\t$ is convex, so is the optimization problem (\ref{SVM-expectile}) and 
by
\cite[Lemma 5.1, Theorem 5.2]{steinwart2008support} there always exits a unique 
$f_{D,\lb}$ that satisfies (\ref{SVM-expectile}). Moreover, 
the solution of $f_{D,\lb}$ is of the form
\begin{align*}
 f_{D,\lb}:= \sum_{i=1}^n (\alpha_i^*-\beta_i^*) K(x_i,\cdot)\,,
\end{align*}
where $\alpha_i^* \geq 0, \beta_i^* \geq 0$ for all $i=i, \ldots, n$, see \cite{farooq2015svm} for further details. 
Learning method of the form (\ref{SVM-expectile}) but with 
different loss functions have attracted many 
theoretical and algorithmic considerations, see for instance 
\cite{wu2006learning, bauer2007regularization, caponnetto2007optimal,
steinwart2009optimal,eberts2013optimal, tacchetti2013gurls} for least square regression,
\cite{steinwart2011estimating, eberts2013optimal} for 
quantile regression and  \cite{hush2006qp,steinwart2011training} for classification with hinge 
loss. In addition, \cite{farooq2015svm} recently  
proposed an algorithm for solving \eqref{SVM-expectile}, that is now a part of 
\cite{Steinwart14b},
and compared  its performance to \textsf{ER-Boost}, see \cite{yang2014nonparametric},
which is another algorithm minimizing an empirical $L_{\tau}$-risk.
The main goal of this article is to complement the empirical findings of \cite{farooq2015svm}
with a detailed statistical analysis.

A typical way to access the quality of an estimator $f_{D}$ is to measure its distance to the 
target function $f_{L_{\tau},P}^{\star}$, e.g.~in terms of 
$\snorm{f_{D}-f_{L_{\tau},P}^{\star}}_{L_2(P_X)}$. For estimators obtained by some 
empirical risk minimization scheme, however, one can hardly ever estimate this $L_2$-norm directly.
Instead, the standard tools of statistical learning theory give bounds on the excess risk
$\frisk{f_{D}}-\orisk$. Therefore, our first goal of this paper is to establish a so-called 
calibration inequality that relates both quantities. To be more precise, we will show in Theorem \ref{self-calibration-inquality} that 
%
%
\begin{equation}\label{function-approx-inequality}
 \snorm{f_{D}-f_{L_{\tau},P}^{\star}}_{L_2(P_X)} \leq c_\t\, \sqrt{\frisk{f_{D}}-\orisk}\,,
\end{equation}
holds for all $f_D\in L_2(P_X)$ and some constant $c_\t$ only depending on $\t$.
In particular, \eqref{function-approx-inequality} provides rates for 
$\snorm{f_{D}-f_{L_{\tau},P}^{\star}}_{L_2(P_X)}$ as soon as we have established rates for 
$\frisk{f_{D}}-\orisk$. 
Furthermore, it is common knowledge in statistical learning theory that bounds on  
$\frisk{f_{D}}-\orisk$ can be improved if so-called variance bounds are available.
We will see in Lemma \ref{Lemma-supremum and variance bound} that \eqref{function-approx-inequality}
leads to an optimal variance bound for $L_\t$ whenever $Y$ is bounded.
Note that both \eqref{function-approx-inequality} and the variance bound are independent of the 
considered expectile estimation method. In fact, both results are key ingredients for the 
statistical analysis of any expectile estimation method based on some form of empirical risk minimization.

%

As already indicated above, however, the main goal of this paper is to provide a statistical analysis of
the SVM-type estimator $f_{D,\lb}$ given by \eqref{SVM-expectile}. Since $2L_{1/2}$ equals the least squares loss,
any statistical analysis of \eqref{SVM-expectile} also provides results for SVMs using the least squares loss. 
The latter have already been extensively investigated in the literature.
 For example,  
learning rates for generic kernels can be found in 
\cite{cucker2002mathematical,de2005model,caponnetto2007optimal,steinwart2009optimal, mendelson2010regularization} 
and the references therein.
Among these articles, only \cite{cucker2002mathematical, steinwart2009optimal, mendelson2010regularization}
obtain learning rates in minimax sense under some specific assumptions. For example, 
\cite{cucker2002mathematical}  assumes that the target function $f_{L_{1/2},P}^{\star} \in H$, while 
\cite{steinwart2009optimal, mendelson2010regularization} establish optimal learning rates 
for the case in which $H$ does not contain the target function. 
In addition, 
 \cite{eberts2013optimal} has recently established (essentially) asymptotically optimal learning
rates for least square SVMs using Gaussian RBF kernels under the assumption that the target function
$f_{L_{1/2},P}^{\star}$ is contained in some Sobolev or Besov space $B_{2,\infty}^\a$ with smoothness index $\a$.
A key ingredient of this work is to 
control the capacity of RKHS $H_{\g}(X)$ for Gaussian RBF kernel $k_{\g}$ 
on the closed unit Euclidean ball $X\subset \R^d$
by 
an entropy number  bound 
\begin{align}
\ e_i(\mathrm{id}:H_{\g}(X) \to l_{\infty}(X) ) 
\leq
c_{p,d}(X) \g^{-\frac d p} i^{-\frac{1}{p}}\,,
\end{align}
see \cite[Theorem 6.27]{steinwart2008support}, which holds for all $\g\in (0,1]$ and  $p\in (0,1]$.
Unfortunately, the constant $c_{p,d}(X)$ derived from 
\cite[Theorem 6.27]{steinwart2008support} depends on $p$ in an unknown manner. As a consequence, 
\cite{eberts2013optimal} were only able to show learning rates of the form 
\begin{align*}
 n^{-\frac{2\a}{2\a+d}+\xi}
\end{align*}
for all $\xi > 0$.
%
To address this issue, we use \cite[Lemma 4.5]{van2009adaptive} to derive 
the following new entropy number bound
%
\begin{align}\label{sec1-improvedEntropy}
e_i (\mathrm{id}: H_{\g}(X) \to l_{\infty}(X)) \leq (3K)^{\frac{1}{p}} 
\left(\frac{d+1}{ep}\right)^{\frac{d+1}{p}} \g^{-\frac{d}{p}} i^{-\frac{1}{p}}\,,
\end{align}
which holds for all $p\in (0,1]$ and $\g\in (0,1]$ and some constant $K$ only depending on $d$.
In other words, we establish an upper bound for $c_{p,d}(X)$ whose dependence on $p$ is explicitly known.
Using this new bound, we are then able to find improved learning rates of the form 
\begin{align*}
(\log n)^{d+1} n^{-\frac{2 \alpha}{2 \alpha+d}}\,.
\end{align*}
Clearly these new rates   replace the nuisance factor $n^\xi$ of \cite{eberts2013optimal}
by some logarithmic term, and up to this logarithmic factor our new rates are minimax optimal,
see \cite{eberts2013optimal} for details.
In addition, our new rates also hold for $\t\neq 1/2$, that is for 
general expectiles.

%

The rest of this paper is organized as follows. In Section \ref{sec-head-propertiesALS}, some properties of 
the ALS loss function 
are established including  the self-calibration inequality and variance bound.
Section \ref{sec-head-OraclelearningRates} presents   oracle inequalities and learning rates for 
\eqref{SVM-expectile}
and Gaussian RBF kernels.
The proofs of our results   can be found in Section \ref{sec-head-proofs}.

\color{black}

\section{Properties of the ALS Loss Function: Self-Calibration and Variance Bounds}\label{sec-head-propertiesALS}
This section contains some properties of the ALS loss function i.e.~convexity, local Lipschitz
continuity, a self-calibration inequality, a supremum bound and a variance bound. 
Throughout this section, we assume that $X$ is an arbitrary, non-empty set equipped with $\sigma$-algebra, 
and $Y \subset \mathbb{R}$ denotes a closed non-empty set. In addition, we assume that $\mathrm{P}$ is the probability 
distribution on $X \times Y$, $P(\cdot|x)$ is a regular conditional probability distribution on $Y$ given 
$x\in X$ and $Q$ is a some distribution on $Y$. Furthermore, $L_{\tau}:Y\times \mathbb{R}\to [0,\infty)$ 
is the ALS loss  defined by \eqref{lossALS} and $f:X\to \mbb{R}$ is a measurable function. 
It is trivial to prove that $L_{\tau}$
is convex in $t$, and  this convexity  ensures that the optimization problem (\ref{SVM-expectile}) is efficiently 
solvable. Moreover, by
\cite[Lemma 2.13]{steinwart2008support} convexity of  $L_{\tau}$ implies convexity of corresponding
risks. In the following, we
present the idea of clipping to restrict the prediction $t$ to the domain $Y = [-M, M]$ 
where $M > 0$, see e.g.~\cite[Definition 2.22]{steinwart2008support}.
\begin{Definition}\label{Definition-clipping}
 We say that a loss $L:Y\times \mathbb{R}\to [0,\infty)$ can be clipped at $M > 0$, if, for all 
 $(y,t) \in Y\times \mathbb{R}$, we have
 \begin{align}\label{clipping}
  L(y, \clip{t}\,) \leq L(y,t)\,,
 \end{align}
where $\clip{t}$ denotes the clipped value of $t$ at $\pm M$, that is
\begin{equation*}
\begin{aligned}
 \clip{t}:=\left\{\begin{array}{ll}
	-M & \hspace*{6ex} \text{if} \hspace*{2ex} t < -M\,,\\ 
	t & \hspace*{6ex} \text{if} \hspace*{2ex} t \in [-M,M]\,,\\   
	M & \hspace*{6ex} \text{if} \hspace*{2ex} t > M\,. 
\end{array}\right.
\end{aligned}
\end{equation*}
Moreover, we say that $L$ can be clipped if $t$ can be clipped at some $M>0$.
\end{Definition}
Recall that this clipping assumption has already been utilized while establishing 
learning rates for SVMs, see for instance \cite{chen2004support, steinwart2006oracle, steinwart2011training} for hinge loss 
and \cite{christmann2007svms, steinwart2011estimating} for 
pinball loss. It is trivial to show by convexity of $L_{\tau}$  together with 
\cite[Lemma 2.23]{steinwart2008support} that  
$L_{\tau}$ can be clipped at $M$  
and has at least one global minimizer in $[-M,M]$. This also implies that 
$\frisk{\clip{f}} \leq \frisk{f}$ for every  $f:X\to \mathbb{R}$. In other words, the clipping operation
potentially reduces the risks.
We therefore bound the risk $\frisk{\clip{f}_{D, \lb}}$ of the clipped decision function 
rather than the risk
$\frisk{f_{D, \lb}}$, which we will see in details in Section \ref{sec-head-OraclelearningRates}. 
From a practical point of view, this means that
the \textit{training} algorithm for 
(\ref{SVM-expectile}) remains unchanged and the \textit{evaluation} of the resulting decision function
requires only a slight change. For further details on algorithmic advantages of clipping for SVMs using 
the hinge loss and the ALS loss, we refer the reader to \cite{steinwart2011training} and \cite{farooq2015svm}
respectively.
It is also observed in \cite{steinwart2008support, steinwart2009optimal, eberts2013optimal} that 
$\inorm{\cdot}$-bounds, see Section \ref{sec-head-OraclelearningRates}, can be made smaller by clipping the decision function for 
some loss functions.

Let us further recall from \cite[Definition 2.18]{steinwart2008support}  that a loss function 
is
%
called locally Lipschitz continuous if for all 
$a\geq 0$ there exists a  constant $c_a$ such that 
\begin{align*}
  \supremum{y\in Y}\abs{L(y,t)-L(y,t')} \leq c_a \abs{t-t'}\,, \hspace*{6ex} t,t'\in [-a,a]\,.
\end{align*}
In the following we denote for a given $a\geq 0$ the smallest such constant $c_a$ by $|L|_{1,a}$.
The following lemma, which we will need for our proofs,  shows that the ALS loss is 
 locally Lipschitz continuous.

\begin{Lemma}\label{Lemma-Lipschitz-constant}
 Let $Y \in [-M,M]$ and $t \in Y$, then the loss function $L_{\tau}: Y \times \mbb{R} \to [0, \infty)$ is
 locally Lipschitz continuous  with Lipschitz constant 
 \begin{equation*}
 \abs{L_{\tau}}_{1,M} = C_\t\,4 M\,,
 \end{equation*}
 where $C_\t:=\max\{\t,1-\t\}$.
\end{Lemma}

For later use note
that $L_{\tau}$  being locally Lipschitz continuous implies that $L_{\tau}$ is also a \textit{Nemitski loss} in the sense of 
\cite[Definition 18]{steinwart2008support}, and by \cite[Lemma 2.13 \mbox{and} 2.19]{steinwart2008support},
this further implies that the corresponding risk $\frisk{f}$ is convex and locally Lipschitz continuous.

Empirical methods of estimating expectile using $L_\t$ loss typically lead to  the function $f_{D}$ 
for which $\frisk{f_{D}}$ is close to $\orisk$ with high probability. 
The convexity of $L_\tau$ then ensures that 
$f_D$  
approximates $f_{L_{\tau},P}^{\star}$ in a weak sense, namely in probability $P_X$,
see \cite[Remark 3.18]{steinwart2007compare}. However, no guarantee on the speed of this convergence can be given,
even if we know the convergence rate of $\frisk{f_{D}} \to \orisk$.
The following theorem addresses this issue by establishing a so-called calibration inequality for the 
excess $L_\tau$-risk.


\begin{Theorem}\label{self-calibration-inquality}
 Let $L_{\tau}$ be the ALS loss function defined by (\ref{lossALS}) and $\mathrm{P}$ be the distribution 
 on $\mathbb{R}$.  Moreover, assume that $f_{L_{\tau},P}^{\star}(x) < \infty$ is the conditional $\tau$-expectile
 for fixed $\tau \in (0,1)$. Then, for all $f:X \to \mbb{R}$, we have
 \begin{equation*}
  C_{\tau}^{-1/2}(\frisk{f}- \orisk)^{1/2} \leq
  \snorm{f-f_{L_{\tau},P}^{\star}}_{L_2(P_X)} 
  \leq c_{\tau}^{-1/2}\, (\frisk{f}- \orisk)^{1/2}\,,
 \end{equation*}
 where $c_\t:=\min\{\t,1-\t\}$ and $C_\t$ is defined in Lemma \ref{Lemma-Lipschitz-constant}.
\end{Theorem}

Note that the calibration inequality, that is the  right-hand side of the inequality above in particular 
ensures that $f_D\to f_{L_{\tau},P}^{\star}$ in $L_2(P_X)$
whenever $\frisk{f_{D}} \to \orisk$. In addition, the convergence rates can be directly translated.
The inequality on the left shows that modulo constants the calibration inequality is sharp.
We will use this left inequality when bounding the approximation error for Gaussian RBF kernels in 
the proof of Theorem \ref{Theorem-approximation_error_function-newEntropy}.

At the end of this section, we present supremum and variance bounds of the $L_{\tau}$-loss.
Like the calibration inequality of Theorem \ref{self-calibration-inquality} these two bounds are 
useful for analyzing the statistical properties  of any $L_\tau$-based empirical risk minimization 
scheme. In Section \ref{sec-head-OraclelearningRates} we will illustrate this when establishing an oracle inequality 
for the SVM-type learning algorithm \eqref{SVM-expectile}.
 
\begin{Lemma}\label{Lemma-supremum and variance bound}
 Let $X \subset \mathbb{R}^d$ be non-empty set, $Y \subset [-M,M]$ be a closed subset where $M > 0$, and
 $\mathrm{P}$ be a distribution on $X \times Y$. Additionally, we assume that 
 $L_{\tau}:Y\times \mathbb{R}\to [0,\infty)$ is the ALS loss   and  $f_{L_{\tau},P}^{\star}(x)$ is the conditional $\tau$-expectile
 for fixed $\tau \in (0,1)$. Then for all $f:X \to [-M,M]$ we have
\begin{itemize}
 \item[i)]$ \inorm{L_{\tau} \circ f -L_{\tau} \circ f_{L_{\tau},P}^{\star}} \leq 4\,C_\t\,M^2\,.$
 \item[ii)] $\mathbb{E}_{P}(L_{\tau} \circ f-L_{\tau} \circ f_{L_{\tau},P}^{\star})^2 
 \leq 16 \,C_\t^2\,c_{\tau}^{-1}\,M^2 (\frisk{f}-\orisk)\,.$
\end{itemize}
\end{Lemma}

\section{Oracle Inequalities and Learning Rates}\label{sec-head-OraclelearningRates}
In this section, we 
first introduce some notions related to kernels. We assume that 
$k: X \times X \to \mathbb{R}$ is a measurable, symmetric and positive definite kernel with 
associated RKHS $H$. Additionally, we assume that $k$ is bounded, that is,  
$\inorm{k}:=\sup_{x\in X} \sqrt{k(x,x)} \leq 1$, which implies
that $H$ consists of bounded functions with $\inorm{f} \leq  \inorm{k}\snorm{f}_{H}$ for all $f \in H$. 
In practice, we often consider SVMs that are equipped with well-known Gaussian RBF kernels for input domain $X \in \mbb{R}^d$, 
see \cite{steinwart2011training, farooq2015svm}.
Recall that the latter are defined by
\begin{align*}
k_{\g}(x,x'):=\exp(-\g^{-2} \snorm{x-x'}_2^2)\,, 
\end{align*}
where $\g$ is called the width parameter that is usually determined in a data dependent way, i.e.~by
cross validation. By \cite[Corollary 4.58]{steinwart2008support} the kernel $k_{\g}$ is universal
on every compact set $X \in \mathbb{R}^n$ and in particular strictly positive definite.
In addition, the RKHS $H_{\g}$  of kernel $k_\g$ is dense in $L_p(\mu)$
for all $p\in [1,\infty)$ and all distributions $\mu$  on $X$, see  \cite[Proposition 4.60]{steinwart2008support}.
	
One requirement to establish learning rates is to control the capacity of RKHS $H$. One way to do this is to 
estimate eigenvalues of a linear operator induced by kernel $k$. To be more precise, 
given a kernel
$k$ and a distribution $\mu$ on $X$, we define the integral operator $T_k:L_2(\mu) \to L_2(\mu)$ by
\begin{equation}\label{integral operator}
 T_kf(\cdot):= \int_X k(x, \cdot)f(x)d\mu(x)
\end{equation}
for $\mu$-almost all $x \in X$. In the following, we assume that $\mu=\mathrm{P}_X$. Recall 
\cite[Theorem 4.27]{steinwart2008support} that $T_k$ is compact, positive, self-adjoint and nuclear, and 
thus has at most countably many non-zero (and non-negative) eigenvalues $\lb_i(T_k)$. Ordering these 
eigenvalues (with geometric multiplicities) and extending the corresponding sequence by zeros, if there are only
finitely many non-zero eigenvalues, we obtain the \textit{extended sequence of eigenvalues} 
$(\lb_i(T_k))_{i \geq 1}$ that satisfies $\sum_{i=1}^{\infty} \lb_i(T_k)< \infty$ 
\cite[Theorem 7.29]{steinwart2008support}.   This summability implies that 
for some constant $a > 1$ and $i \geq 1$, we have $ \lb_i(T_k) \leq a i^{-1}$. 
By \cite{steinwart2009optimal}, this eigenvalues assumption
can converge even faster to zero, that is, for $p \in (0,1)$, we have
\begin{align}\label{eigne-value-assumption}
 \lb_i(T_k) \leq a i^{-\frac{1}{p}}, \hspace*{6ex} i \geq 1.
\end{align}
It turns out that the speed of convergence of $\lb_i(T_k)$ influences 
learning rates for SVMs. For instance,  \cite{blanchard2008statistical} used (\ref{eigne-value-assumption}) 
to establish learning rates for SVMs using hinge loss and 
\cite{caponnetto2007optimal,mendelson2010regularization} for SVMs 
using least square loss. 

Another way to control the capacity of RKHS $H$ is based on the concept of 
\textit{covering numbers} or the inverse of covering numbers, namely, \textit{entropy numbers}. 
To recall the latter, see \cite[Definition A.5.26]{steinwart2008support}, let  
$T:E \to F$ be a bounded, linear operator between the Banach spaces $E$ and $F$, and $i \geq 1$ be an integer.
Then the $i$-th (dyadic) entropy number of $T$ is defined by
\begin{align*}
 e_i(T):=\inf \left\{\eps > 0: \exists x_1, \ldots, x_{2^{i-1}} \,\mathrm{such\, that}\,\,
 TB_E \subset \cup_{j=1}^{2^{i-1}}(x_j+\eps B_F)  \right\}\,.
\end{align*}
In the Hilbert space case, the eigenvalues and entropy number decay are closely related. For example, 
\cite{steinwart2009oracle} showed that (\ref{eigne-value-assumption}) is 
equivalent (modulo a constant only depending on $p$) to
\begin{align}\label{general-entropy}
 e_i (\mathrm{id}: H \to L_2(P_X))\leq \sqrt{a}i^{-\frac{1}{2p}}\,, \hspace*{6ex} i \geq 1\,,
\end{align}
It is further shown in \cite{steinwart2009oracle} that 
(\ref{general-entropy}) implies a bound on average entropy numbers, that is, for empirical distribution
 associated to the data set $D_X:=(x_1, \cdots, x_n)\in X^n$, the average entropy number is
 \begin{align*}
  \mathbb{E}_{D_X \sim P_X^n} e_i(\mathrm{id}:H \to L_2(P_X) ) \leq a i^{-\frac{1}{2p}},\hspace*{6ex} i \geq 1\,,
 \end{align*}
which is used in \cite[Theorem 7.24]{steinwart2008support} to establish the general oracle inequality for SVMs.
A bound of the form (\ref{general-entropy}) was also established by \cite[Theorem 6.27]{steinwart2008support} 
for Gaussian RBF kernels and certain distributions $P_X$ having unbounded support. 
To be more precise, let $X \subset \mbb{R}^d$ be a closed unit Euclidean ball. Then for all 
$\g \in (0,1]$ and $p\in (0,1)$, there exists a constant $c_{p,d}(X)$ such that
\begin{align}\label{entropy-RBF-old}
\ e_i(\mathrm{id}:H_{\g}(X) \to l_{\infty}(X) ) 
\leq
c_{p,d}(X) \g^{-\frac{d}{p}} i^{-\frac{1}{p}}\,,
\end{align}
which has been 
used by \cite{eberts2013optimal} to establish leaning rates for least square SVMs. 
Note that the constant $c_{p,d}(X)$ depends on $p$ in an unknown manner. To address this 
issue, we use 
\cite[lemma 4.5]{van2009adaptive} and derive an improved entropy number bound in the following 
theorem by establishing an upper bound for $c_{p,d}(X)$ whose dependence on $p$ is explicitly known.
We will further see in Corollary \ref{Corollary-lr_for_entropy_2} that this improved bound leads us 
to achieve better learning rates than the one obtained by \cite{eberts2013optimal}.
\begin{Theorem}\label{sec3-theorem-newEntropyNumber}
Let $X = \R^d$ be a closed Euclidean ball. Then there exists a constant $K > 0$, 
such that, for all $p \in (0,1)$,  $\g \in (0,1]$ and $i \geq 1$, we have
\begin{align}\label{sec3-eq-newEntropyNumber}
e_i (\mathrm{id}: H_{\g}(X) \to l_{\infty}(X)) \leq (3K)^{\frac{1}{p}} 
\left(\frac{d+1}{ep}\right)^{\frac{d+1}{p}} \g^{-\frac{d}{p}} i^{-\frac{1}{p}}
\end{align}
\end{Theorem}
	

Another requirement for establishing learning rates is to bound 
the  \textit{approximation error function} considering RKHS $H_\g$ for Gaussian RBF kernel $k_\g$. If the 
distribution $P$ is such that $\orisk < \infty$, then the approximation error function  $\mcal{A}:[0, \infty) \to [0, \infty)$
is defined by 
\begin{equation}\label{approximation error function-newEntropy}
 \mathcal{A}(\lb):=\infimum{f\in {H_\g}} \lb \snorm{f}_{H_\g}^2 + \frisk{f}-\orisk\,.
\end{equation}
For $\lb >0$, the approximation error function 
$\mathcal{A}(\lb)$ quantifies how well an infinite sample $L_2$-SVM with RKHS $H_\g$, that is,  $\lb \hnorm{f}^2 + \frisk{f}$ approximates
the optimal risk $\orisk$. By \cite[Lemma 5.15]{steinwart2008support}, one can show that $\lim_{\lb \to 0} \mathcal{A}(\lb)=0$ if
$H_\g$ is dense in $L_2(P_X)$.
In general, however, the speed of convergence can not be faster than $O(\lb)$ and this rate is achieved, 
if and only if, there exists an $f \in H_\g$ such that $\frisk{f}=\orisk$, 
see \cite[Lemma 5.18]{steinwart2008support}. 
%
%

In order to bound $\mcal{A}(\lb)$,
we first need to know one important feature of the target function $f_{L_{\tau},P}^{\star}$, namely, the \textit{regularity} 
which, roughly speaking, measures the smoothness of the target function. Different function spaces norms e.g.~H{\"o}lder norms, Besov norms or
Triebel-Lizorkin norms can be used to capture this regularity. 
In this work, following \cite{eberts2013optimal,meister2016optimal}, 
we assume that the target function $f_{L_{\tau},P}^{\star}$ is in  a Sobolev or a
 Besov space. Recall \cite[Definition 5.1]{tartar2007introduction} and \cite[Definition 3.1 and 3.2]{adams2003sobolev} 
that for any integer $k \geq 0$, 
$1\leq p \leq \infty$ and a subset $\Omega \subset \R^d$ with non-empty interior, the Sobolev space $W_p^k(\Omega)$ 
of order $k$ is defined by
\begin{align*}
 W_p^k(\Omega):=
 \{f\in L_p(\Omega): D^{(\a)}f \in L_p(\Omega) \text{ exists for all } \a \in \mbb{N}_0^d \text{ with } \abs{\a} \leq k \}\,,
\end{align*}
with the norm
\begin{equation*}
\begin{aligned}
\snorm{f}_{W_p^k(\Omega)}:=\left\{\begin{array}{ll}
	\Big(\sum_{\abs{\a}\leq k}\snorm{D^{(\a)}f}_{L_p(\Omega)}^p\Big)^{\frac{1}{p}}\,, & \hspace*{6ex} \text{if} \hspace*{2ex} p \in [1, \infty)\,,\\ 
	\max\sum_{\abs{\a}\leq k}\snorm{D^{(\a)}f}_{L_{\infty}(\Omega)}\,,     & \hspace*{6ex} \text{if} \hspace*{2ex} p=\infty\,,
\end{array}\right.
\end{aligned}
\end{equation*}
where $D^{(\a)}$ is the $\a$-th weak
partial derivative for multi-index $\a=(\a_1, \ldots, \a_d)\in \mbb{N}_0^d$ of modulus $\abs{\a}=\abs{\a_1}+\cdots + \abs{\a_d}$.
In other words, the Sobolev space is the space of functions with sufficiently many derivatives and equipped 
with a norm that measures both the size and the regularity of the contained functions. 
Note that $W_p^k(\Omega)$ is a Banach space, see \cite[Lemma 5.2]{tartar2007introduction}. Moreover, 
by \cite[Theorem 3.6]{adams2003sobolev}, $W_p^k(\Omega)$ is separable if $p \in [1, \infty)$, and is 
uniformly convex and reflexive if $p \in (1,\infty)$. Furthermore, for $p=2$, 
$W_2^k(\Omega)$ is a separable Hilbert space  that we denote by $H_k(\Omega)$.
Despite the underlined advantages, Sobolev spaces can not be immediately applied when $\a$ is non-integral or when 
$p < 1$, however, the smoothness spaces for these extended parameters are also needed when engaging 
nonlinear approximation. This shortcoming of Sobolev spaces is covered by Besov spaces that bring together 
all functions for which the modulus of smoothness have a common behavior. Let us first recall 
 \cite[Section 2]{devore1993besov} and
\cite[Section 2]{devore1988interpolation}
that for a subset $\Omega \subset \R^d$ with non-empty interior,
a function $f:\Omega \to \R$ with $f\in L_p(\Omega)$ for all 
$p\in (0,\infty]$ and $s\in \mbb{N}$, the modulus of smoothness of order $s$ of a  function $f$ is defined by
\begin{align*}
 w_{s,L_p(\Omega)}(f,t)=\supremum{\snorm{h}_2 \leq t}\snorm{\triangle_h^s (f,\cdot)}_{L_p(\Omega)}\,,\hspace*{4ex} t\geq 0\,,
\end{align*}
where the $s$-th difference $\triangle_h^s (f,\cdot)$ given by
\begin{equation*}
\begin{aligned}
\triangle_h^s (f,x, \Omega):=\left\{\begin{array}{ll}
	\sum_{i=0}^s \binom{r}{i} (-1)^{r-i}f(x+ih)& \hspace*{6ex} 
			\text{if}\hspace*{2ex} x, x+h, \ldots, x+sh \in \Omega\,,\\ 
		0,     & \hspace*{6ex} \text{otherwise}\,,
\end{array}\right.
\end{aligned}
\end{equation*}
for $h\in \mbb{R}^d$, is used to measure the smoothness. Note that 
$w_{s,L_p(\Omega)}(f,t) \to 0$ as $t\to 0$, which means that the faster this convergence to 0 the smoother is 
$f$. For more details on properties of the modulus of smoothness, we refer
the reader to \cite[Chapter 4.2]{nikol2012approximation}.
Now for $0 < p,q \leq \infty$, $\a > 0$, $s:= \lfloor \a \rfloor +1$,
the Besov space $B_{p,q}^{\a}(\Omega)$ based on modulus of smoothness for domain $\Omega \subset \R^d$, 
see for instance \cite[Section 4.5]{devore1998nonlinear}, \cite[Chapter 4.3]{nikol2012approximation} 
and \cite[Section 2]{devore1993besov}, is defined by
\begin{align*}
 B_{p,q}^{\a}(\Omega):=\{f\in L_p(\Omega):\abs{f}_{B_{p,q}^{\a}(\Omega)}< \infty\}\,,
\end{align*}
where the semi-norm $\abs{\cdot}_{B_{p,q}^{\a}(\Omega)}$ is given by
\begin{align*}
 \abs{f}_{B_{p,q}^{\a}(\Omega)}:=
 \Big(\int_0^{\infty} (t^{-\a} w_{s,L_p(\Omega)}(f,t) )^q \frac{dt}{t}\Big)^{\frac{1}{q}}\,, \hspace*{4ex} q\in (0,\infty)\,,
\end{align*}
and for $q=\infty$, the semi-norm $\abs{\cdot}_{B_{p,q}^{\a}(\Omega)}$ is defined by
\begin{align*}
 \abs{f}_{B_{p,q}^{\a}(\Omega)}:= \supremum{t>0} (t^{-\a} w_{s,L_p(\Omega)}(f,t))\,.
\end{align*}
In other words, Besov spaces are collections of functions $f$ with common smoothness. For more general 
definition of Besov-like spaces, we refer to \cite[Section 4.1]{meister2016optimal}. Note that 
 $\snorm{f}_{B_{p,q}^{\a}(\Omega)}:=\snorm{f}_{L_p(\Omega)}+\abs{f}_{B_{p,q}^{\a}(\Omega)}$ is the norm of 
$B_{p,q}^{\a}(\Omega)$, see e.g.~\cite[Section 2]{devore1993besov} 
and \cite[Section 2]{devore1988interpolation}.
Furthermore, for $p > 1$ different values of $s > \a$ give equivalent norms of 
$B_{p,q}^{\a}(\Omega)$, which 
remains true for $p<1$, see \cite[Section 2]{devore1993besov}. It is well known, 
see e.g~\cite[Section 4.1]{nikol2012approximation}, that $W_p^s(\Omega) \subset B_{p,\infty}^s(\Omega)$ for all
$1\leq p \leq \infty$, $p \neq 2$, where for $p=q=2$ the Besov space is the same as the Sobolev space.

 

In the next step, we find a function $f_0 \in H_{\g}$  such 
that both the regularization term $\lb \snorm{f_0}_{H_{\g}}^2$ and the excess risk 
$\frisk{f_0}-\orisk$ are small. For this, 
we define the function $K_{\g}:\mathbb{R}^d\to \mathbb{R}$, see \cite{eberts2013optimal}, by 

\begin{equation}
 K_{\g}(x):=\sum_{j=1}^r \binom rj (-1)^{1-j}\frac{1}{j^d}\Big(\frac{2}{\g^2 \pi}\Big)^{\frac{d}{2}}
 \exp\Big(-\frac{2\snorm{x}_2^2}{j^2 \g^2}\Big)\,,
\end{equation}
for all $r\in \mathbb{N}$, $\g > 0$ and $x\in \mathbb{R}^d$. Additionally, we assume that there exists
a function 
$f_{L_{\tau},P}^{\star}:\mathbb{R}^d \to \mathbb{R}$ satisfies  
$f_{L_{\tau},P}^{\star}\in  L_2(\mathbb{R}^d)\cap  L_{\infty} (\mathbb{R}^d) $
and $\frisk{f_{L_{\tau},P}^{\star}}=\orisk$.
Then $f_0$ is defined by

\begin{equation*}
 f_0(x):= K * f_{L_{\tau},P}^{\star}(x)
 :=\int_{\mathbb{R^d}} K(x-t)f_{L_{\tau},P}^{\star}(t)dt\,, \hspace*{6ex} x \in \mathbb{R^d}\,.
\end{equation*}
With these preparation, we now establish an upper bound for the approximate error function $\mcal{A}(\lb)$.

%

\begin{Theorem}\label{Theorem-approximation_error_function-newEntropy}
 Let $L_{\tau}$ be the ALS loss  defined by (\ref{lossALS}), $\mathrm{P}$ be the
 probability distribution
 on $\mathbb{R}^d \times Y$, and $P_X$ be the marginal distribution of $P$ onto $\R^d$  
 such that $X:=\mathrm{supp}\, \mathrm{P}_X$ and
 $P_X(\partial X)=0$. 
 Moreover, assume that the conditional $\tau$-expectile $f_{L_{\tau},P}^{\star}$ satisfies 
 $f_{L_{\tau},P}^{\star} \in L_2 (\mathbb{R}^d) \cap L_{\infty} (\mathbb{R}^d)$ as well as 
 $f_{L_{\tau},P}^{\star} \in B_{2,\infty}^{\alpha}(P_X)$ for some $\alpha \geq 1$. In addition, assume that $k_{\g}$ 
is the Gaussian RBF kernel over $X$ with  
 associated RKHS  $H_{\g}$. Then for all $\g \in (0,1]$ and $\lb > 0$, we have
 \begin{align*}
  \hnorm{f_0}^2 +\frisk{f_0}-\orisk \leq C_1 \lb \g^{-d}+C_{\tau,s} \g^{2\alpha}\,,
 \end{align*}
 where $C_{\tau,s}>0$
 is a constant depending on $s$ and $\tau$, and the constant $C_{1}>0$.
\end{Theorem}

Clearly, 
the upper bound of the approximation error function in Theorem \ref{Theorem-approximation_error_function-newEntropy}
depends on the regularization parameter $\lb$, the kernel width $\g$, and the smoothness parameter
$\alpha$ of the target function $f_{L_{\tau},P}^{\star}$.
Note that in order to shrink the right-hand side we need to let $\g\to 0$.
However, this would let the first term go to infinity unless we simultaneously let $\lb \to 0$ with a
sufficient speed.
Now using \cite[Theorem 7.24]{steinwart2008support} together with Lemma 
\ref{Lemma-supremum and variance bound}, Theorem \ref{Theorem-approximation_error_function-newEntropy} 
and the entropy number bound (\ref{sec3-eq-newEntropyNumber}), we establish oracle 
inequality of SVMs for 
$L_{\t}$ in the following theorem.

\begin{Theorem}\label{Theorem-final_orale_inequlity-newEntropy}
 Consider the assumptions of Theorem \ref{Theorem-approximation_error_function-newEntropy} and additionally assume that
 $Y:=[-M,M]$ for $M \geq 1$. 
 Then, for all 
 $n\geq 1, \varrho \geq 1, \g \in (0,1)$ and $\lb \in (0,e^{-2}]$, the SVM using the RKHS $H_{\g}$ and the
 ALS  loss function $L_{\tau}$ satisfies
 \begin{align}\label{sec3-eq-finalOracle}
  \lb \snorm{f_{D, \lb,\g}}_{H_{\g}}^2 + \frisk{\clip{f}_{D, \lb,\g}}-\orisk
  &\leq
  CM^2 \Big(\lb \g^{-d} + \g^{2\alpha}+(\log \lb^{-1})^{d+1}\,n^{-1}\g^{-d} + n^{-1}\varrho\Big)\,,
  \end{align}
 with probability $P^n$ not less than $1-3 e^{-\varrho}$. Here $C>0$ is some constant independent of $p, \lb, \g, n$ and 
 $\varrho$.
\end{Theorem}

It is well known that there exists a relationship between Sobolev spaces and 
the scale of Besov spaces, that is,
$B_{p,u}^{\a}(\mathbb{R}^d)\hookrightarrow W_p^{\a}(\mathbb{R}^d) \hookrightarrow B_{p,v}^{\a}(\mathbb{R}^d)$,
whenever $1 \leq u \leq \min\{p,2\}$ and $\max\{p,2\} \leq v \leq \infty$, see for instance \cite[p.25 and p.44]{edmunds2008function}. 
In particular, for $p=u=v=2$, 
we have $W_2^{\a}(\mathbb{R}^d) = B_{2,2}^{\a}(\mathbb{R}^d)$ with equivalent norms. In addition, 
by \cite[p.7]{eberts2013optimal} we have $B_{p,q}^{\a}(\mathbb{R}^d) \subset B_{p,q}^{\a}(P_X) $. Thus, 
Theorem \ref{Theorem-final_orale_inequlity-newEntropy} also holds for  decision functions 
$f_{L_{\tau},P}^{\star}:\mathbb{R}^d \to \mathbb{R}$ with 
$f_{L_{\tau},P}^{\star} \in L_2(\mathbb{R}^d) \cap  L_{\infty}(\mathbb{R}^d)$ and 
$f_{L_{\tau},P}^{\star} \in W_2^{\a}(\mathbb{R}^d)$.

By assuming some suitable values for $\lb$ and $\g$ that depends on data size $n$, the smoothness
parameter $\a$, and the  dimension $d$, we obtain   learning rates for 
learning problem (\ref{SVM-expectile}) in the following corollary.

\begin{Corollary}\label{Corollary-lr_for_entropy_2}
 Under the assumptions of Theorem \ref{Theorem-final_orale_inequlity-newEntropy} and  with
 \begin{align*}
  \lb_{n}&=c_1 n^{-1}\,,\\
  \g_{n}&=c_2 n^{-\frac{1}{2\alpha + d}}\,,
  \end{align*}
where $c_1 > 0$ and $c_2 >0$ are user specified constants, we have, for all $n \geq 1$ and $\varrho \geq 1$,
\begin{equation}\label{lr_entropy_2}
 \frisk{\clip{f}_{D, \lb, \g}}-\orisk
 \leq C M^2 \varrho (\log n)^{d+1} n^{-\frac{2 \alpha}{2 \alpha+d}}
\end{equation}
with probability $P^n$ not less than $1-3e^{-\varrho}$.
\end{Corollary}

Note that learning rates in Corollary \ref{Corollary-lr_for_entropy_2} depend on the choice of 
$\lb_n$ and $\g_n$, where the kernel width $\g_n$ requires knowing $\alpha$ which, in practice, 
is not available.
However, 
\cite[Chapter 7.4]{steinwart2008support}, \cite{steinwart2009optimal}, \cite{eberts2013optimal} and 
\cite{steinwart2011estimating} 
showed that one can achieve the same learning rates adaptively, i.e.~without knowing $\alpha$.
Let us recall \cite[Definition 6.28]{steinwart2008support} that describes a method to select $\lb$ and $\g$,
which in some sense is a simplification of the cross-validation method.
\begin{Definition}
 Let $H_{\g}$ be a RKHS over $X$ and $\Lambda:=(\Lambda_n)$ and $\Gamma:=(\Gamma_n)$ be the sequences of 
 finite subsets $\Lambda_n, \Gamma_n \subset (0,1]$. Given a data set 
 $D:= ((x_1, y_1), \ldots, (x_n, y_n)) \in (X \times \mathbb{R})^n$, we define
 \begin{align*}
  D_1 &:= ((x_1, y_1), \ldots, (x_m, y_m)) \\
  D_2 &:= ((x_{m+1}, y_{m+1}), \ldots, (x_n, y_n))\,,
 \end{align*}
where $m = \lfloor \frac{n}{2} \rfloor +1$ and $n \geq 4$. Then use $D_1$ as a training set to compute
the SVM decision function
\begin{equation*}
 f_{D_1, \lb, \g}:= \arg \underset{f \in H_{\g}}{\min} \lb \snorm{f}_{H_{\g}}^2
 + \mathcal{R}_{L_{\tau}, D_1}(f), \hspace*{6ex} (\lb, \g) \in (\Lambda_n, \Gamma_n)\,,
\end{equation*}
and use $D_2$ to determine $(\lb, \g)$ by choosing 
$(\lb_{D_2}, \g_{D_2}) \in (\Lambda_n, \Gamma_n)$ such that
\begin{equation*}
 \mathcal{R}_{L_{\tau}, D_2} (\clip{f}_{D_1, \lb_{D_2}, \g_{D_2}}) 
  = \underset{(\lb, \g) \in (\Lambda_n, \Gamma_n)}{\min}\, \mathcal{R}_{L_{\tau}, D_2} (\clip{f}_{D_1, \lb, \g})\,. 
\end{equation*}
Every learning method that produce the resulting decision functions 
$\clip{f}_{D_1, \lb_{D_2}, \g_{D_2}}$ is called a training validation SVM with respect to 
$(\Lambda, \Gamma)$.
\end{Definition}

In the next Theorem, we use this training-validation SVM (TV-SVM) approach for suitable candidate sets 
$\Lb_n:= (\lb_1, \ldots, \lb_r)$ and $\G_n:=(\g_1, \ldots, \g_s)$ with $\lb_r=\g_s=1$,  
and establish learning
rates similar to (\ref{lr_entropy_2}).

\begin{Theorem}\label{Theorem-TV-SVM learning rates-newEntropy}
 With the assumptions of Theorem \ref{Theorem-final_orale_inequlity-newEntropy}, let $\Lb:=(\Lambda_n)$
 and $\Gamma:=(\Gamma_n)$ be the sequences of finite subsets $\Lb_n, \G_n \subset (0,1]$ such that 
 $\Lb_n$ is an $n^{-1}$-net of $(0,1]$ and $\G_n$ is an $n^{-\frac{1}{2 \alpha+d}}$-net of $(0,1]$
 with polynomially growing cardinalities $|\Lambda_n|$ and $|\Gamma_n|$ in $n$.
  Then for all $\varrho \geq 1$, 
 the TV-SVM produce $f_{D_1, \lb_{D_2}, \g_{D_2}}$  that satisfies
 \begin{equation*}
  P^n \Big(\frisk{\clip{f}_{D_1, \lb_{D_2}, \g_{D_2}}} -\orisk 
  \leq
  C M^2 \varrho (\log n)^{(d+1)} \,n^{-\frac{2\alpha}{2\alpha+d}}\Big) \geq 1-3e^{-\varrho}
 \end{equation*}
 where $C > 0$ is a constant independent of $n$ and $\varrho$.
\end{Theorem}

So far we have only considered the case of bounded noise  with known bounds, that is, $Y \in [-M, M]$
where $M > 0$ is known.
In practice, $M$ is usually unknown and in this situation, one  can still achieve the same 
 learning rates by simply 
increasing $M$ slowly. However, more interesting is the case of unbounded noise. 
In the following we treat this case for distributions for which  
there exist  constants $c\geq 1$ and 
$l > 0$ such that
\begin{equation}\label{Unbounded noise-entropy2}
 P(\{(x,y) \in X \times Y : |y| \leq c \varrho^l\}) \geq 1-e^{-\varrho}
\end{equation}
for all $\varrho > 1$. 
In other words, the tails of the response variable $Y$ decay sufficiently fast.
It is shown in \cite{eberts2013optimal} by examples that such an assumption is realistic. 
For instance, if $P(.|x) \sim N(\mu(x),1)$, the assumption \eqref{Unbounded noise-entropy2} is
satisfied for $l=\frac{1}{2}$, see \cite[Example 3.7]{eberts2013optimal}, 
and for the case where $P(.|x)$ has the density whose tails decay like 
$e^{-\abs{t}}$, the assumption 
\eqref{Unbounded noise-entropy2} holds for   $l=1$, see \cite[Example 3.8]{eberts2013optimal}.

%

With this additional assumption, we present learning rates for the case of unbounded noise in the 
following theorem.

\begin{Theorem}\label{Theorem-unbounded noise learning rates-newEntropy}
 Let $Y \subset \mathbb{R}$ and $\mathrm{P}$ be a probability distribution on 
 $\mathbb{R}^d \times Y$ such that 
 $ X:=\mathrm{supp}\,\mathrm{P}_X \subset B_{l_2^d}$. Moreover, assume that the $\tau$-expectile 
 $f_{L_{\tau},P}^{\star}$ satisfies $f_{L_{\tau},P}^{\star}(x) \in [-1,1]$ for $\mathrm{P}_X$-almost 
 all $x \in X$, and both
 $f_{L_{\tau},P}^{\star} \in L_2 (\mathbb{R}^d) \cap L_{\infty} (\mathbb{R}^d)$ and
 $f_{L_{\tau},P}^{\star} \in B_{2,\infty}^{\alpha}(P_X)$ for some $\alpha \geq 1$.
 In addition, assume that
 (\ref{Unbounded noise-entropy2}) holds for all $\varrho \geq 1$. We define
 \begin{align*}
  \lb_n & = c_1 n^{-1}\\
  \g_n &=c_2 n^{-\frac{1}{2\alpha+d}}\,,
 \end{align*}
where $c_1>0$ and $c_2>0$ are user-specified constants. Moreover, for some fixed 
$\hat{\varrho} \geq 1$ and $n \geq 3$ we define $\varrho:=\hat{\varrho}+\ln n$ and $M_n:=2c\varrho^l$. Furthermore, we 
consider the SVM that clips decision function $f_{D, \lb_n, \g_n}$ at $M_n$ after training. 
Then there exists a $C > 0$ independent of $n$, $p$ and $\hat{\varrho}$ such that
\begin{equation}\label{lr-unbounded case}
 \lb_n \snorm{f_{D, \lb_n, \g_n}}_{H_{\g_n}}^2 
  + \frisk{\clip{f}_{D, \lb_n, \g_n}}-\orisk
  \leq
  C \hat{\varrho}^{2l+1} (\log n)^{2l+d+1} n^{-\frac{2\alpha}{2 \alpha+d}}
\end{equation}
holds with probability $P^n$ not less than $1-2e^{-\hat{\varrho}}$.
\end{Theorem}

Note that the assumption (\ref{Unbounded noise-entropy2}) on the tail of the distribution does not 
influence learning rates achieved in the Corollary \ref{Corollary-lr_for_entropy_2}. Furthermore, 
we can also achieve same rates adaptively using TV-SVM approach considered in Theorem \ref{Theorem-TV-SVM learning rates-newEntropy}
provided that we have upper bound of the unknown parameter $l$, which depends on the distribution $P$, see
 \cite{eberts2013optimal} where this dependency is explained with some examples. 

 Let us now compare our results with the oracle inequalities and learning rates established by 
\cite{eberts2013optimal} for least square SVMs. This comparison is justifiable because 
a) the least square loss is a special case of $L_\t$-loss for $\t=0.5$, 
b) the target function $f_{L_\t,P}^{\star}$ is assumed to be in the Sobolev or Besov space similar to \cite{eberts2013optimal}, and 
c) the supremum and the variance bounds for $L_\t$ with $\tau=0.5$ are 
the same as the ones  used by \cite{eberts2013optimal}.
Furthermore, recall that  \cite{eberts2013optimal} used the 
entropy number bounds \eqref{entropy-RBF-old} to control the capacity of the RKHS $H_\g$  which contains a constant   $c_{p,d}(X)$   depending on $p$ in an unknown manner.  As a result, they obtained 
a leading constant  $C$ in their oracle inequality, see \cite[Theorem 3.1]{eberts2013optimal} for which no  upper bound 
can   be determined explicitly.
We cope this problem by establishing
an improved entropy number bound \eqref{sec3-eq-newEntropyNumber} which not only provides the upper bound for $c_{p,d}(X)$ but also helps to 
determine the value of the constant $C$ in the oracle inequality \eqref{sec3-eq-finalOracle} explicitly.
As a consequence we can improve their learning rates of the form 
$ n^{-\frac{2\a}{2\a+d}+\xi}\,$,
where $\xi > 0$, by
\begin{align}\label{sec3-eq-learningRates}
 (\log n)^{d+1} \,n^{-\frac{2\a}{2\a +d}}\,.
\end{align}
In other words,  the nuisance parameter $n^\xi$ from \cite{eberts2013optimal} is replaced by the logarithmic term $(\log n)^{d+1}$. Moreover,
our learning rates, up to this logarithmic term, are minimax optimal, see e.g.~the discussion in \cite{eberts2013optimal}.
Finally note that  
unlike \cite{eberts2013optimal}
we have not only established learning rates for the least squares case $\t=0.5$ but 
actually for all $\t\in (0,1)$.

\section{Proofs}\label{sec-head-proofs}
\subsection{\textit{Proofs of Section \ref{sec-head-propertiesALS}}}

\begin{proofof}{Lemma \ref{Lemma-Lipschitz-constant}}
We define $\psi:\R \to \R$ by
\begin{equation*}
\begin{aligned}
\psi(r):=\left\{\begin{array}{ll}
	(1-\tau) r^2 \,, & \hspace*{6ex} \text{if} \hspace*{2ex} r < 0\,,\\ 
	\tau r^2\,,     & \hspace*{6ex} \text{if} \hspace*{2ex} r \geqslant 0\,.
\end{array}\right.
\end{aligned}
\end{equation*}
Clearly, $\psi$ is convex and thus \cite[Lemma A.6.5]{steinwart2008support} shows that $\psi$ is 
locally Lipschitz continuous. Moreover, we have
\begin{align*}
 \abs{L(Y,\cdot)}_{1,M} &= \abs{\psi(Y-\cdot)}_{1,M}
			=\supremum{t\in[-M,M]} \abs{\psi'(y-t)}_{1,M}
			\leq \max\{\t,1-\t\} \supremum{t\in[M,-M]}\abs{2(y-t)}
			\leq C_{\t}\,4M\,,
\end{align*}
where $C_\t:=\max\{\t,1-\t\}$. A simple consideration shows that this estimate is also sharp.
%
\end{proofof}

In order to prove Theorem  \ref{self-calibration-inquality}
recall 
 that the risk $\frisk{f}$ in \eqref{sec1-eq-risk} uses regular conditional probability
$P(y|x)$, which enable us to
computed $\frisk{f}$ by treating the \textit{inner} and the
\textit{outer} integrals separately. Following 
\cite[Definition 3.3, Definition 3.4]{steinwart2008support}, we therefore use \textit{inner $L_{\tau}$-risks}
as a key ingredient for establishing self-calibration inequalities.

\begin{Definition}\label{Definition-inner_L_risk}
 Let $L_{\tau}: Y \times \mathbb{R} \to [0, \infty)$ be the ALS loss function defined by 
 (\ref{lossALS}) and $Q$ be a distribution on $Y=[-M,M]$. Then the \textit{inner $L_{\tau}$-risks} of 
 $Q$ are defined by
 \begin{equation*}
 \mathcal{C}_{L_{\tau},Q}(t):= \int_{Y}L_{\tau}(y,t)dQ(y)\,, \hspace*{6ex} t \in \mathbb{R}\,,
\end{equation*}
and the \textit{minimal inner $L_{\tau}$-risk} is
\begin{equation*}
 \mathcal{C}_{L_{\tau},Q}^{\star}:=\underset{t \in \mathbb{R}}{\inf}\, \mathcal{C}_{L_{\tau},Q}(t)\,.
\end{equation*}
\end{Definition}

In the latter definition, the \textit{inner risks} $\mathcal{C}_{L_{\tau},Q}(\cdot)$ for 
a suitable classes of distributions $Q$ on $Y$ are considered as a template for $\mathcal{C}_{L_{\tau},P(\cdot|x)}(\cdot)$.
From this, 
we immediately can obtain the risk of function $f$, i.e.~ 
$\frisk{f}=\int_X \mathcal{C}_{L_{\tau},P(\cdot|x)}(f(x) dP_X(x)$.
Moreover,  
by \cite[Lemma 3.4]{steinwart2008support}, the optimal risk $\orisk$ can be obtained by minimizing 
the \textit{inner $L_{\tau}$-risks}, i.e.
~$\orisk= \int_{X} \mathcal{C}_{L_{\tau},P(\cdotp|x)}^{\star} dP_X(x)$. 
consequently, the \textit{excess $L_{\tau}$-risk}, when $\orisk< \infty$, is 
obtained by
\begin{equation} \label{excess-risk-from-inner-risk}
 \frisk{f}-\orisk= \int_{X} \mathcal{C}_{L_{\tau},P(\cdotp|x)}(f(x))-\mathcal{C}_{L_{\tau},P(\cdotp|x)}^{\star} dP_X(x)\,.
\end{equation}
Besides some technical advantages, this approach makes the analysis rather independent of the 
specific distribution $\mathrm{P}$. In the following theorem, we use this approach and 
establish the lower and the 
upper bound of excess inner 
$L_{\tau}$-risks.

\begin{Theorem}\label{Theorem-lower-upper-bound-excess-inner-risk}
 Let $L_{\tau}$ be the ALS loss function defined by (\ref{lossALS})
 and $Q$ be a distribution on  $\mathbb{R}$ with $\mathcal{C}_{L_{\tau},Q}^{\star} < \infty$.
 For a fixed $\tau \in (0,1)$ and for all $t \in \mathbb{R}$, we have
\begin{equation}\label{lower-upper-bound-excess-inner-risk}
c_{\tau}(t-t^{\star})^2 
 \leq
 \mathcal{C}_{L_{\tau},Q}(t)-\mathcal{C}_{L_{\tau},Q}^{\star} 
 \leq 
 C_{\tau}(t-t^{\star})^2\,,
\end{equation}
where $c_{\tau}:=\min\{\tau, 1-\tau\}$ and $C_{\tau}$ is defined in Lemma \ref{Lemma-Lipschitz-constant}.
\end{Theorem}
\begin{proofof}{Theorem \ref{Theorem-lower-upper-bound-excess-inner-risk}}
 Let us fix $\tau \in (0,1)$. Then for a distribution $Q$ on $\mbb{R}$  satisfies
 $\mcal{C}_{L_{\tau},Q}^{\star}< \infty$,
 the  $\tau$-expectile $t^{\star}$, according to \cite{newey1987asymmetric}, is the only solution of 
\begin{equation}\label{Newey-expectile-expression}
 \tau \int_{y \geq t^{\star}} (y-t^{\star})dQ(y) = (1-\tau) \int_{y < t^{\star}}(t^{\star}-y)dQ(y)\,.
\end{equation}
Let us now compute the excess inner risks of $L_{\tau}$ with respect to $Q$. To this end, we 
fix a $t \geq t^{\star}$. Then we have
 \begin{align*}
  \int_{y < t}(y-t)^2 dQ(y) &=\int_{y < t}(y-t^{\star}+t^{\star}-t)^2 dQ(y)\\
		&= \int_{y < t} (y-t^{\star})^2 dQ(y) + 2(t^{\star} -t) \int_{y < t} (y-t^{\star})dQ(y) + (t^{\star} -t)^2 Q((-\infty,t))\\
		&=\int_{y < t^{\star}}(y-t^{\star})^2 dQ(y)+\int_{t^{\star} \leq y < t}(y-t^{\star})^2 dQ(y) +(t^{\star}-t)^2 Q((-\infty, t))\\
		& \quad +2(t^{\star}-t)\int_{y < t^{\star}}(y-t^{\star}) dQ(y)+2(t^{\star}-t)\int_{t^{\star} \leq y < t}(y-t^{\star}) dQ(y)\,,
 \end{align*}
and
\begin{align*}
  \int_{y \geq t}(y-t)^2 dQ(y)&=\int_{y \geq t^{\star}}(y-t^{\star})^2 dQ(y)-\int_{t^{\star} \leq y < t}(y-t^{\star})^2 dQ(y) +(t^{\star}-t)^2 Q([t, \infty))\\
		& \quad +2(t^{\star}-t)\int_{y \geq t^{\star}}(y-t^{\star}) dQ(y)- 2(t^{\star}-t)\int_{t^{\star} \leq y < t}(y-t^{\star}) dQ(y)\,.
\end{align*}
By Definition \ref{Definition-inner_L_risk} and using (\ref{Newey-expectile-expression}), we obtain
\begin{align*}
 \mathcal{C}_{L_{\tau},Q}(t)&= (1-\tau) \int_{y < t}(y-t)^2 dQ(y)+\tau \int_{y \geq t}(y-t)^2 dQ(y)\\
			&= \tau \int_{y < t^{\star}}(y-t^{\star})^2 dQ(y) + (1-\tau) \int_{y \geq t^{\star}}(y-t^{\star})^2 dQ(y)\\
			& \quad + 2(t^{\star}-t) \Big(\tau \int_{y < t^{\star}}(y-t^{\star}) dQ(y) + (1-\tau) \int_{y \geq t^{\star}}(y-t^{\star}) dQ(y) \Big)\\
			& \quad  + (t^{\star}-t)^2 (1-\tau) Q((-\infty, t))+(t^{\star}-t)^2 \tau Q([t, \infty))\\
			& \quad +(1-2\tau) \int_{t^{\star} \leq y < t}(y-t^{\star})^2 dQ(y) + 2(1-2\tau) \int_{t^{\star} \leq y < t}(y-t^{\star}) dQ(y)\\
		     &= \mathcal{C}_{L_{\tau},Q}(t^{\star})+(t^{\star}-t)^2 (1-\tau) Q((-\infty, t))+(t^{\star}-t)^2 \tau Q([t, \infty))\\
			&\quad  +(1-2\tau)\int_{t^{\star} \leq y < t}(y-t^{\star})^2+ 2(t^{\star}-t)(y-t^{\star}) dQ(y)\,,
\end{align*}
and this leads to the following excess inner $L_{\tau}$-risk 
\begin{align}
&\mathcal{C}_{L_{\tau},Q}(t)-\mathcal{C}_{L_{\tau},Q}(t^{\star})\nonumber\\
&=(t^{\star}-t)^2 (1-\tau)Q((-\infty, t^{\star}))+ (t^{\star}-t)^2 (1-\tau)Q([t^{\star}, t))+(t^{\star}-t)^2 \tau Q([t, \infty))\nonumber\\
			& \quad +(1-2\tau)\int_{t^{\star} \leq y < t}(y-t^{\star})^2+ 2(t^{\star}-t)(y-t^{\star}) dQ(y)\nonumber\\
&= (t^{\star}-t)^2 \Big((1-\tau)Q((-\infty, t^{\star}))+\tau Q([t, \infty))\Big)-\tau\int_{t^{\star} \leq y < t}(y-t^{\star})^2+ 2(t^{\star}-t)(y-t^{\star}) dQ(y)\nonumber\\
			&\quad + (t^{\star}-t)^2 (1-\tau)Q([t^{\star}, t)) +(1-\tau)\int_{t^{\star} \leq y < t}(y-t^{\star})^2+ 2(t^{\star}-t)(y-t^{\star}) dQ(y)\nonumber\\
&= (t^{\star}-t)^2 \Big((1-\tau)Q((-\infty, t^{\star}))+\tau Q([t, \infty))\Big)-\tau\int_{t^{\star} \leq y < t}(y-t^{\star})(y+t^{\star}-2t) dQ(y)\nonumber\\
			&\quad +(1-\tau)\int_{t^{\star} \leq y < t}(y-t^{\star})^2+ 2(t^{\star}-t)(y-t^{\star})+(t^{\star}-t)^2 dQ(y)\nonumber\\
&=(t^{\star}-t)^2 \Big((1-\tau)Q((-\infty, t^{\star}))+\tau Q([t, \infty))\Big)+ \tau\int_{t^{\star} \leq y < t}(y-t^{\star})(2t-t^{\star}-y) dQ(y)\nonumber \\
		& \quad +(1-\tau)\int_{t^{\star} \leq y < t}(y-t)^2 dQ(y)\,.\label{proof-EIR when t >=t^*}	
\end{align}
Let us define $c_{\tau}:=\min\{\tau, 1-\tau\}$, then (\ref{proof-EIR when t >=t^*}) leads to the following 
lower bound of excess inner $L_{\tau}$-risk when $t \geq t^{\star}$:
\begin{align}
&\mathcal{C}_{L_{\tau},Q}(t)-\mathcal{C}_{L_{\tau},Q}(t^{\star})\nonumber\\
& \geq c_{\tau}(t^{\star}-t)^2 \Big(Q((-\infty, t^{\star}))+Q([t, \infty))\Big)
	+ c_{\tau} \int_{t^{\star} \leq y < t} (y-t^{\star})(2t-t^{\star}-y)+(y-t)^2 dQ(y)\nonumber\\
& =c_{\tau}(t^{\star}-t)^2 \Big(Q((-\infty, t^{\star}))+Q([t, \infty))\Big)
	+ c_{\tau} \int_{t^{\star} \leq y < t} (t^{\star})^2+2tt^{\star}+t^2 dQ(y)\nonumber\\
& = c_{\tau}(t^{\star}-t)^2 \Big(Q((-\infty, t^{\star}))+Q([t, \infty))\Big)
	+ c_{\tau}(t^{\star}-t)^2 Q([t^{\star},t))\nonumber\\
& = c_{\tau}(t^{\star}-t)^2\,.\label{proof-lb_EIR when t >=t^*}
\end{align}
Likewise, the excess inner $L_{\tau}$-risk when $t < t^{\star}$ is
\begin{equation}\label{proof-EIR when t<t^*}
 \begin{aligned}
  \mathcal{C}_{L_{\tau},Q}(t)-\mathcal{C}_{L_{\tau},Q}(t^{\star})
  &=(t^{\star}-t)^2 \Big((1-\tau)Q((-\infty, t)+\tau) Q([t^{\star},\infty))\Big)+\tau \int_{t \leq y < t^{\star}}(y-t)^2 dQ(y)\\
	    &\quad +(1-\tau)\int_{t \leq y < t^{\star}}(t^{\star}-y)(y+t^{\star}-2t) dQ(y)\,,
 \end{aligned}
\end{equation}
that also leads to the lower bound (\ref{proof-lb_EIR when t >=t^*}).
Now, for the proof of upper bound of the excess inner $L_{\tau}$-risks, 
we define $C_{\tau}:=\max\{\tau, 1-\tau\}$. 
Then (\ref{proof-EIR when t >=t^*}) leads to the following upper bound of 
excess inner $L_{\tau}$-risks when $t \geq t^{\star}$: 
\begin{align}
&\mathcal{C}_{L_{\tau},Q}(t)-\mathcal{C}_{L_{\tau},Q}(t^{\star})\nonumber\\
& \leq C_{\tau}(t^{\star}-t)^2 \Big(Q((-\infty, t^{\star}))+Q([t, \infty))\Big)
	+ C_{\tau} \int_{t^{\star} \leq y < t} \big((y-t^{\star})(2t-t^{\star}-y)+(y-t)^2\big)dQ(y)\nonumber\\
& = C_{\tau}(t^{\star}-t)^2\,.\label{proof-ub_EIR when t >=t^*}
\end{align}
Analogously, for the case of $t < t^{\star}$, (\ref{proof-EIR when t<t^*}) also leads to the upper bound 
(\ref{proof-ub_EIR when t >=t^*}) for excess inner $L_{\tau}$-risks.
\end{proofof}

\begin{proofof}{Theorem \ref{self-calibration-inquality}}
For a fixed $x\in X$, we write $t:=f(x)$ and $t^{\star}:=f_{L_{\tau},P}^{\star}(x)$.
By Theorem \ref{Theorem-lower-upper-bound-excess-inner-risk}, for $Q:=P(\cdot|x)$, we then immediately obtain
%
\begin{align*}
  C_{\tau}^{-1}(\mathcal{C}_{L_{\tau},P(\cdotp|x)}(f(x))-\mathcal{C}_{L_{\tau},P(\cdotp|x)}^{\star})
  \leq \abs{f(x)- f_{L_{\tau},P}^{\star}(x)}^2 
  \leq c_{\tau}^{-1} \,(\mathcal{C}_{L_{\tau},P(\cdotp|x)}(f(x))-\mathcal{C}_{L_{\tau},P(\cdotp|x)}^{\star})\,.
\end{align*}
Integrating with respect to $P_X$ leads to the assertion.
\end{proofof}

\begin{proofof}{Lemma \ref{Lemma-supremum and variance bound}}
i) Since $L_{\tau}$ can be clipped at $M$ and the conditional $\tau$-expectile satisfies 
$f_{L_{\tau},P}^{\star}(x) \in [-M,M]$ almost surely. Then 
\begin{align*}
 \inorm{L_{\tau}(y, f(x))-L_{\tau}(y, f_{L_{\tau},P}^{\star}(x))}
 &\leq \max\{\tau, 1-\tau\} \supremum{y,t \in [-M,M]}(y-t)^2\\
 &= C_\t\,4M^2\,,
 \end{align*}
for all $f:X \to [-M, M]$ and all $(x,y) \in X \times Y$. \\
ii) Using the locally Lipschitz continuity of the loss  $L_{\tau}$ and Theorem 
\ref{self-calibration-inquality}, we obtain
\begin{align*}
\mathbb{E}_P(L_{\tau} \circ f - L_{\tau} \circ f^{\star}_{\tau,P})^2 
&\leq \abs{L_{\tau}}_{1,M}^2 \,\,\mathbb{E}_{P_X}\abs{f-f_{\tau,P}^{\star}}^2\\
&\leq 16 c_{\tau}^{-1}C_\t^2\,M^2\, (\frisk{f}-\orisk)\,.
\end{align*}
\end{proofof}

\subsection{\textit{Proofs of Section \ref{sec-head-OraclelearningRates}}}
\begin{proofof}{Theorem \ref{sec3-theorem-newEntropyNumber}}
 By \cite[Lemma 4.5]{van2009adaptive}, the $\inorm{\cdot}$-log covering numbers of unit ball $B_{\g}(X)$ of 
 the Gaussian RKHS $H_{\g}(X)$ for all $\g \in (0,1)$ and $\varepsilon \in (0, \frac{1}{2})$
 satisfy 
 \begin{equation}\label{appnd-proof-coveringNumber}
  \mathcal{H}_{\infty}(B_\g(X), \varepsilon) 
  \leq K \left(\log \frac{1}{\varepsilon}\right)^{d+1} \g^{-d}\,,
 \end{equation}
where $K>0$ is a constant depending only on $d$. From this, 
we conclude that
\begin{equation*}
 \sup_{\varepsilon \in (0, \frac{1}{2})} \varepsilon^{p} \mathcal{H}_{\infty}(B_\g(X), \varepsilon)
 \leq 
 K  \g^{-d} \sup_{\varepsilon \in (0, \frac{1}{2})} \varepsilon^{p} \left(\log \frac{1}{\varepsilon}\right)^{d+1}\,.
\end{equation*}
Let $h(\varepsilon):=\varepsilon^{p} \left(\log \frac{1}{\varepsilon}\right)^{d+1}$. In order to obtain 
the optimal value of $h(\varepsilon)$, we differentiate it with respect to $\varepsilon$
\begin{align*}
 \frac{d h(\varepsilon)}{d \varepsilon}= p \varepsilon^{p-1} \left(\log \frac{1}{\varepsilon}\right)^{d+1}
 - \varepsilon^{p}(d+1) \left(\log \frac{1}{\varepsilon}\right)^{d}\frac{1}{\varepsilon}\,,
\end{align*}
and set $\frac{d h(\varepsilon)}{d \varepsilon}=0$ which gives
\begin{align*}
 \log \frac{1}{\varepsilon}&=\frac{d+1}{p}\\
 \varepsilon^*&=\frac{1}{e^{\frac{d+1}{p}}}\,.
\end{align*}
By plugging $\varepsilon^*$ into $h(\varepsilon)$, we obtain
\begin{align*}
 h(\varepsilon^*)=\left(\frac{d+1}{ep}\right)^{d+1}\,,
\end{align*}
and consequently, $\inorm{\cdot}$-log covering numbers (\ref{appnd-proof-coveringNumber}) are
\begin{align*}
  \mathcal{H}_{\infty}(B_\g(X), \varepsilon) 
  &\leq 
  K \left(\frac{d+1}{ep}\right)^{d+1} \g^{-d} \varepsilon^{-p}
  =\bgparenth{\frac{a^{\frac{1}{p}}}{\varepsilon}}^p\,,
\end{align*}
where $a:= K \left(\frac{d+1}{ep}\right)^{d+1} \g^{-d} $.
Now, by inverse implication of \cite[Lemma 6.21]{steinwart2008support}, 
see also \cite[Exercise 6.8]{steinwart2008support},
the bound on entropy number of the Gaussian RBF kernel is
\begin{align*}
e_i(\mathrm{id}: \mathcal{H}_{\g}(X) \to l_{\infty}(X)) \leq (3a)^{\frac{1}{p}} i^{-\frac{1}{p}}
= (3K)^{\frac{1}{p}}\left(\frac{d+1}{ep}\right)^{\frac{d+1}{p}}\g^{-\frac{d}{p}}\,i^{-\frac{1}{p}} \,,
\end{align*}
for all $i \geq 1$, $\g\in (0,1)$.
\end{proofof}

\begin{proofof}{Theorem \ref{Theorem-approximation_error_function-newEntropy}} 
The assumption $f_{L_{\tau},P}^{\star} \in L_2(\mathbb{R}^d)$ and \cite[Theorem 2.3]{eberts2013optimal} immediately
yield that $f_0:=K * f_{L_{\tau},P}^{\star} \in H_{\g}$, i.e. $f_0$ is contained in RKHS $H_{\g}$. Furthermore,
\cite[Theorem 2.3]{eberts2013optimal} leads to the following upper bound of the regularization term
\begin{equation*}
 \snorm{f_0}_{H_{\g}} = \snorm{K * f_{L_{\tau},P}^{\star}}_{H_{\g}}\leq (\g \sqrt{\pi})^{-\frac{d}{2}} (2^s-1)
 \snorm{f_{L_{\tau},P}^{\star}}_{L_2(\mathbb{R}^d)}.
\end{equation*}
In the next step, we bound the excess risk. By \cite[Theorem 2.2]{eberts2013optimal}, the upper bound for $L_2(P_X)$-distance 
between $f_0$ and $f_{L_{\tau},P}^{\star}$ is
\begin{align}\label{excess-risk-upper-bound-newEntropy}
\snorm{f_0-f_{L_{\tau},P}^{\star}}_{L_2(P_X)}^2
	= \snorm{K * f_{L_{\tau},P}^{\star}-f_{L_{\tau},P}^{\star}}_{L_2(P_X)}^2
	\leq C_{s,2} \snorm{g}_{L_2(\mbb{R}^d)} c^2 \g^{2\a}\,,
\end{align}
where  
$C_{s, 2}:=:=\sum_{i=0}^{\lceil 2s \rceil} \binom{\lceil 2s \rceil}{i} (2d)^{\frac{i}{2}} \prod_{j=1}^{i}(j-\frac{1}{2})^{\frac{1}{2}}$,
see \cite[p.27]{eberts2013optimal},
is constant only depending on $s$ and $g \in L_2(\mbb{R}^d)$ is the Lebesgue density.
Now using Theorem 
 \ref{Theorem-lower-upper-bound-excess-inner-risk} together with (\ref{excess-risk-upper-bound-newEntropy}), we obtain
 \begin{align*}
\frisk{f_0}-\orisk 
	& \leq C_{\tau}\, \snorm{f_0-f_{L_{\tau},P}^{\star}}_{L_2(P_X)}^2=C_{\tau,s} \g^{2\alpha}\,,
\end{align*}
where $C_{\tau,s}:= c^2\,C_{\tau}\,C_{s,2}\, \snorm{g}_{L_2(\mbb{R}^d)} $.
With these results, we finally obtain
\begin{align*}
\underset{f\in {H_{\g}}}{\inf} \lb \snorm{f}_{H_{\g}}^2 + \frisk{f}-\orisk\,
&\leq \lb \snorm{f_0}_{H_{\g}}^2 + \frisk{f_0}-\orisk\,,\\
&\leq  C_1 \lb \g^{-d}+ C_{\tau,s} \g^{2\alpha}\,,
\end{align*}
where $C_1:=(\sqrt{\pi})^{-d} (2^r-1)^2
 \snorm{f_{L_{\tau},P}^{\star}}_{L_2(\mathbb{R}^d)}^2$.
\end{proofof}

In order to prove the main oracle inequality given in Theorem \ref{Theorem-final_orale_inequlity-newEntropy},
we need the following lemma.
\begin{Lemma}\label{apndx-lem-convexDerivative}
The function $h:(0,\frac{1}{2}] \to \mathbb{R}$ defined by
 \begin{align*}
  h(p):=\left(\frac{\sqrt{2}-1}{\sqrt{2}-2^{\frac{2p-1}{2p}}}\right)^p\,,
 \end{align*}
is convex. Moreover, we have $\sup_{p\in (0,\frac{1}{2}]}h(p)=1$.
\end{Lemma}

\begin{proof}
By considering the linear transformation $t:=2p$, it is suffices to show that the function
$g:(0,1]\to \mathbb{R}$ defined by
 \begin{align*}
  g(t):=\left(\frac{\sqrt{2}-1}{\sqrt{2}-2^{1-\frac{1}{t}}}\right)^{\frac{t}{2}}\,,
 \end{align*}
is convex. To solve the latter, we first compute the first and second derivative  of 
$g(t)$ with respect to $t$, that is:
\begin{align*}
 g'(t)=\frac{1}{2}\left(\frac{\sqrt{2}-1}{\sqrt{2}-2^{1-\frac{1}{t}}}\right)^{\frac{t}{2}}
	\left(\log\left(\frac{\sqrt{2}-1}{\sqrt{2}-2^{1-\frac{1}{t}}}\right)+\frac{2^{1-\frac{1}{t}}\log 2}{t \big(\sqrt{2}-2^{1-\frac{1}{t}}\big)}\right)\,,
\end{align*}
and
\begin{align}\label{apndx-lem-convexDerivative-result}
 g''(t)&= \left(\frac{\sqrt{2}-1}{\sqrt{2}-2^{1-\frac{1}{t}}}\right)^{\frac{t}{2}}
		\left(\frac{1}{2}\log\left(\frac{\sqrt{2}-1}{\sqrt{2}-2^{1-\frac{1}{t}}}\right)+\frac{2^{1-\frac{1}{t}}\log 2}{2t \big(\sqrt{2}-2^{1-\frac{1}{t}}\big)}\right)^2\nonumber\\
		&\quad +\left(\frac{\sqrt{2}-1}{\sqrt{2}-2^{1-\frac{1}{t}}}\right)^{\frac{t}{2}}
			\left(\frac{\big(2^{1-\frac{1}{t}}\big)^2 (\log2)^2}{2 t^3 \big(\sqrt{2}-2^{1-\frac{1}{t}}\big)^2}+\frac{2^{1-\frac{1}{t}} (\log2)^2}{2 t^3 \big(\sqrt{2}-2^{1-\frac{1}{t}}\big)}\right)
\end{align}
Since $t\in (0,1]$, it is not hard to see that all terms in $g''(t)$ are strictly positive. Thus 
$g''(t)>0$ and hence $g(t)$ is convex. 
Furthermore, by convexity of $g(t)$, it is easy to find that
\begin{align*}
 \sup_{t \in (0,1]}g(t)=\max\{\lim_{t \to 0}g(t), g(1)\}=1.
\end{align*}
\end{proof}

\begin{proofof}{Theorem \ref{Theorem-final_orale_inequlity-newEntropy}}
The assumption $f_{L_{\tau},P}^{\star} \in L_{\infty}(\mathbb{R}^d)$
 and \cite[Theorem 2.3]{eberts2013optimal} yield that
 \begin{equation*}
  |K * f_{L_{\tau},P}^{\star}(x)|\leq (2^s-1)\snorm{f_{L_{\tau},P}^{\star}}_{L_{\infty}(\mathbb{R}^d)}\,,
 \end{equation*}
holds for all $x\in X$. This implies that, for all $(x,y)\in X\times Y$, we have
\begin{align*}
 L_{\tau}(y,K * f_{L_{\tau},P}^{\star}(x))&\leq (M + \inorm{K * f_{L_{\tau},P}^{\star}})^2\\
				    &\leq 4 (M + 2^s\snorm{f_{L_{\tau},P}^{\star}}_{L_{\infty}(\mathbb{R}^d)})^2:=B_0\,,
\end{align*}
and hence we conclude that $B_0 \geq 4M^2$. Now, by plugging the result of Theorem \ref{Theorem-approximation_error_function-newEntropy}
together with $a=(3K)^{\frac{1}{2p}} \Big(\frac{d+1}{e p}\Big)^{\frac{d+1}{2p}}$ from 
Theorem \ref{sec3-theorem-newEntropyNumber} 
and $V=16 c_{\tau}^{-1}\, M^2$ from Lemma \ref{Lemma-supremum and variance bound}, into
\cite[Theorem 7.23]{steinwart2008support}, we obtain
\begin{align}\label{proof-oracle inequality-first resutl}
 \lb \snorm{f_{D, \lb,\g}}_{H_{\g}}^2 + \frisk{\clip{f}_{D, \lb, \g}}-\orisk
  &\leq
  9\,C_1 \lb \g^{-d}+ 9\,C_{\tau,s} \g^{2\alpha}
  +3 K(p)\, K \Big(\frac{d+1}{e}\Big)^{d+1}\frac{\g^{-d}}{p^{d+1}\lb^p n}\nonumber\\
  &\quad + (3456 M^2\, C_\t^2\,c_{\tau}^{-1}+60(M + 2^s\snorm{f_{L_{\tau},P}^{\star}}_{L_{\infty}(\mathbb{R}^d)})^2)\frac{\varrho}{n}\,,\nonumber\\
  & \leq 9\,C_1 \lb \g^{-d}+ 9\,C_{\tau,s} \g^{2\alpha}+ C_d\,K(p)\,\frac{\g^{-d}}{p^{d+1}\lb^p n}
  +C_2 \frac{\varrho}{n}\,,
\end{align}
where $C_1$ and $C_{\tau,s}$ are from Theorem \ref{Theorem-approximation_error_function-newEntropy}, 
$K(p)$ is a constant from  \cite[Theorem 7.23]{steinwart2008support} that depends on $p$,
$ C_2:= 3456\, M^2 \,C_\t^2\, c_{\tau}^{-1}+60(M + 2^s\snorm{f_{L_{\tau},P}^{\star}}_{L_{\infty}(\mathbb{R}^d)})^2$,
and $C_d:=3 K \Big(\frac{d+1}{e}\Big)^{d+1}$ is a constant only depending on $d$.
Let us assume that  $p:=\frac{1}{\log \lb^{-1}}$. Since $\lb \leq e^{-2}$ and $\lb^p=e^{-1}$,thus
(\ref{proof-oracle inequality-first resutl}) becomes 
\begin{align}\label{proof-oracle-inequality-second-result}
 \lb \snorm{f_{D, \lb,\g}}_{H}^2 + \frisk{\clip{f}_{D, \lb, \g}}-\orisk
&\leq 9\,C_1 \lb \g^{-d}+ 9\,C_{\tau,s} \g^{2\alpha}+ C_d \, e\,K(p)\,(\log \lb^{-1})^{d+1}\, \frac{ \g^{-d}}{n}
  +C_2 \frac{\varrho}{n}
\end{align}
We now consider the constant $K(p)$ in more detail. To this end, by using the Lipschitz constant $\abs{L_\tau}_{1,M}=4M$
from  Lemma \ref{Lemma-Lipschitz-constant} and the supremum bound $B=4M^2$ from  Lemma \ref{Lemma-supremum and variance bound}
, the value of $K(p)$ is, see \cite[Theorem 7.23]{steinwart2008support}: 
\begin{align}\label{proof-constant K}
 K:&=3 \max\{30 \cdot 2^p\,C_1(p)\abs{L_\tau}_{1,M}^p \,V^{\frac{1-p}{2}},
	30 \cdot (120)^p\,C_2^{1+p}(p)\abs{L_\tau}_{1,M}^{2p}\,B^{1-p},B\}\nonumber\\
   &=3 \max \{120 \cdot 2^{p}\,M \,C_\t\,c_{\tau}^{(p-1)/2}\,C_1(p), 120 \cdot (480)^p\,C_\t^{1+p}\, M^2\, C_2^{1+p}(p), C_\t\,4\,M^2\}\,,	
\end{align} 
where the constants $C_1(p)$ and $C_2(p)$ are derived in the proof of \cite[Theorem 7.16]{steinwart2008support}, that is
\begin{align*}
 C_1(p):=\frac{2\sqrt{\ln 256}C_p^p}{(\sqrt{2}-1)(1-p)2^{p/2}} \hspace*{4ex} \text{and} \hspace*{4ex}
 C_2(p):=\left(\frac{8\sqrt{\ln16}C_p^p}{(\sqrt{2}-1)(1-p)4^p}\right)^{\frac{2}{1+p}}\,,
\end{align*}
and by \cite[Lemma 7.15]{steinwart2008support}, we have
\begin{align*}
 C_p:=\frac{\sqrt{2}-1}{\sqrt{2}-2^{\frac{2p-1}{2p}}}.\frac{1-p}{p}.
\end{align*}
Here we are interested to bound $K(p)$ for $p \in (0, \frac{1}{2}]$. For this, we first need to bound the constants $C_1(p)$ and $C_2(p)$.
We start with $C_p$ and obtain the following bound for $p\in (0, \frac{1}{2}]$. 
\begin{align*}
C_p^p	&= \Big(\frac{\sqrt{2}-1}{\sqrt{2}-2^{\frac{2p-1}{2p}}}\Big)^p  \Big(\frac{1-p}{p}\Big)^p 
	\leq e \Max{p \in (0,\frac{1}{2}]} \Big(\frac{\sqrt{2}-1}{\sqrt{2}-2^{\frac{2p-1}{2p}}}\Big)^p=e\,,
\end{align*}
where we used $\Big(\frac{1-p}{p}\Big)^p=\Big(\frac{1}{p}-1\Big)^p \leq e$ for all $p\in (0, \frac{1}{2}]$, and 
Lemma \ref{apndx-lem-convexDerivative}.
Now the bound for $C_1(p)$ is the following:
\begin{align*}
 C_1(p) \leq \Max{p\in (0,\frac{1}{2}]}\frac{2 \sqrt{\ln 256}\,C_p^p}{(\sqrt{2}-1)(1-p)2^{p/2}}
	 \leq  \frac{4\, e\,\sqrt{\ln 256}}{\sqrt{2}-1}\,\Max{p\in (0,\frac{1}{2}]} \,\frac{1}{2^{p/2}}
	 \leq 46 \,e\,. 
\end{align*}
Analogously, the bound for the constant $C_2(p)$ is:
\begin{align*}
 C_2^{1+p}(p)\leq \Max{p\in (0, \frac{1}{2}]} \left(\frac{8\sqrt{\ln16}\,C_p^p}{(\sqrt{2}-1)(1-p)4^p}\right)^2
	      \leq \frac{256\, e^2 \ln (16)}{(\sqrt{2}-1)^2} \Max{p\in (0, \frac{1}{2}]}\frac{1}{4^{2p}}
	      \leq 1035\, e^2\,.						
\end{align*}
By plugging  $C_1(p)$ and $C_2(p)$ into (\ref{proof-constant K}), we thus obtain 
\begin{align*}
 K & \leq 3\,\max\{8 \cdot 10^4\,C_\t\, c_{\tau}^{-1/2}\, e \, M, 
		  9 \cdot 10^7\, C_\t \,e^2 M^2,C_\t\,4M^2 \}\\
   & \leq 3 \cdot 10^8\,C_\t \,c_{\tau}^{-1/2}\,e^2\,M^2\,,
  \end{align*}
and by plugging this result into (\ref{proof-oracle-inequality-second-result}), we obtain
\begin{align*}
  \lb \snorm{f_{D, \lb,\g}}_{H}^2 + \frisk{\clip{f}_{D, \lb, \g}}-\orisk
  \leq
  C M^2\Big( \lb \g^{-d}+  \g^{2\alpha}+ (\log \lb^{-1})^{d+1} \g^{-d} n^{-1}
  +\varrho\,n^{-1}\Big)\,,
\end{align*}
where $C$ is a constant 
independent of $p, \lb, \g, n$ and $\varrho$.
\end{proofof}

\begin{proofof}{Corollary \ref{Corollary-lr_for_entropy_2}}
For all $n \geq 1$, Theorem \ref{Theorem-final_orale_inequlity-newEntropy} yields
 \begin{align*}
  \lb \snorm{f_{D, \lb,\g}}_{H_{\g}}^2 + \frisk{\clip{f}_{D, \lb,\g}}-\orisk
  &\leq
  c M^2 (\log \lb^{-1})^{d+1} \Big(\lb_n \g_n^{-d} + \g_n^{2\a} + n^{-1} \g_n^{-d} + n^{-1} \varrho\Big)
\end{align*}
with probability $P^n$ not less than $1-3e^{-\varrho}$ and a constant $c > 0$. Using the sequences 
$\lb_n=c_1 n^{-1}$ and $ \g_n=c_2 n^{-\frac{1}{2\alpha+d}}$, we obtain 
\begin{align*}
 \lb \snorm{f_{D, \lb,\g}}_{H_{\g}}^2 + \frisk{\clip{f}_{D, \lb,\g}}-\orisk
  &\leq
  C M^2(\log n)^{d+1} \Big((c_1 c_2^{-d}+c_2^{2\a}+c_2^{-d})n^{-\frac{2\a}{2\a+d}}+n^{-1} \varrho\Big)\\
  &  \leq \tilde{C}  M^2\varrho (\log n)^{d+1} n^{-\frac{2\a}{2\a+d}}\,,
\end{align*}
where the positive constant $\tilde{C}:=C(c_1 c_2^{-d}+c_2^{2\a}+c_2^{-d}+1)$ is independent of $p$.
\end{proofof}

Before we can proof the Theorem 
\ref{Theorem-TV-SVM learning rates-newEntropy},
we need the following technical lemma.

\begin{Lemma}\label{Lemma-TV-SVM-RHS}
Let $c \geq 3$, $n\geq 3$ be a constant, $\Lb_n \subset (0,1]$ be a finite set such that there 
exists a $\lb_i \in \Lb_n$
with $\frac{1}{c}n^{-1}\leq \lb_i \leq c n^{-1}$. Moreover assume that $\delta_n \geq 0$ and 
$\G_n \subset (0,1]$ is a finite $\d_n$-net of
 $(0,1]$. Then for $d> 0$ and $\a > 0$ we have
 \begin{align*}
 \infimum{(\lb,\g)\in \Lb_n \times \G_n} (\lb \g^{-d} + \g ^{2\a} + (\log \lb^{-1})^{d+1}\g^{-d} n^{-1}  )
 \leq
 c (\log n)^{d+1}\Big(n^{-\frac{2\a}{2\a +d}}+ \d_n^{2\a}\Big)\,,
 \end{align*}
where $c$ is a constant independent of $n, \d_n, \Lb_n, \G_n$.
\end{Lemma}

\begin{proof}
Let us assume that $\Lb_n = \{\lb_1, \ldots, \lb_r\}$ and  $\G_n = \{\g_1, \ldots, \g_s\}$, 
and $\lb_{i-1} < \lb_i$ for all $i= 2, \ldots, r$ and $\g_{j-1} < \g_j$ for all 
$j=2, \ldots, s$. We thus obtain
\begin{align}\label{proof-lemma-data-dependent-LR}
\infimum{(\lb,\g)\in \Lb_n \times \G_n} \Bparenth{\lb \g^{-d}+\g^{2\a}+\frac{(\log \lb^{-1})^{d+1}}{\g^{d} n}}
 &\leq \infimum{\g \in \G_n} \Bparenth{\lb_i \g^{-d}+\g^{2\a}+\frac{(\log \lb_i^{-1})^{d+1}}{\g^{d} n}}\nonumber\\
 &\leq \infimum{\g \in \G_n} \Bparenth{c\g^{-d} n^{-1}+\g^{2\a}+(\log c + \log n)^{d+1}\g^{-d} n^{-1}}\nonumber\\
 & \leq \big(c + (2\log c)^{d+1}\big)\,(\log n)^{d+1}\,\infimum{\g\in \G_n} \Bparenth{\g^{-d} n^{-1}+\g^{2\a}}\nonumber\\
 &\leq \tilde{c}\,(\log n)^{d+1}\infimum{\g\in \G_n} \Bparenth{\g^{-d} n^{-1}+\g^{2\a}}\,,
\end{align}
where $\tilde{c}:=c + (2\log c)^{d+1}$. It is not hard to see that the function $\g \mapsto \g^{-d} n^{-1}+\g^{2\a}$ is optimal at 
$\g^{*}:= c_1 n^{-\frac{1}{2\a+d}}$, where $c_1 > 0$ is a constant only depends on $\a$ and $d$. Furthermore, with $\g_0=0$, we see that
$\g_j-\g_{j-1} \leq 2 \delta_n$ for all $j=1,\ldots,s$. In addition, there exits an index 
$j\in \{1, \ldots, s\}$ such that $\g_{j-1}\leq \g_n^*\leq \g_j$. Consequently, we have 
$\g_n^* \leq \g_j\leq \g_n^*+2\delta_n$. Using this result in (\ref{proof-lemma-data-dependent-LR}), we obtain
\begin{align*}
\infimum{(\lb,\g)\in \Lb_n \times \G_n} \Bparenth{\lb \g^{-d}+\g^{2\a}+\frac{(\log \lb^{-1})^{d+1}}{\g^{d} n}}
&  \leq \tilde{c}\,(\log n)^{d+1}\,\Bparenth{\g_j^{-d} n^{-1}+\g_j^{2\a}}\\
& \leq \tilde{c}\,(\log n)^{d+1}\,\Bparenth{(\g_n^*)^{-d} n^{-1}+(\g_n^*+2\delta_n)^{2\a}}\\
& \leq \tilde{c}\, (\log n)^{d+1} \Bparenth{(\g_n^*)^{-d} n^{-1}+ c_{\a}(\g_n^*)^{2\a}+ c_{\a}\delta_n^{2\a}}\\
& \leq \tilde{c}_{\a}\, (\log n)^{d+1} \Bparenth{( c_1 n^{-\frac{1}{2\a+d}})^{-d} n^{-1}+( c_1 n^{-\frac{1}{2\a+d}})^{2\a}+\delta_n^{2\a}}\\
& \leq c\, (\log n)^{d+1} \Bparenth{ n^{-\frac{2\a}{2\a+d}}+\delta_n^{2\a} }\,,
\end{align*}
where $c:=\tilde{c}_{\a}(c_1^{-d}+c_1^{2\a})$ is a constant.
\end{proof}

\begin{proofof}{Theorem \ref{Theorem-TV-SVM learning rates-newEntropy}}
The proof of this theorem is the literal repetition of the proof of \cite[Theorem 3.6 ]{eberts2013optimal}, however, we
present here for the sake of completeness.
Let us define $m:= \lfloor \frac{n}{2}\rfloor +1 \geq \frac{n}{2} $, then for all $(\lb,\g) \in \Lb_n \times \G_n$, 
Theorem \ref{Theorem-final_orale_inequlity-newEntropy} yields
 \begin{align*}
  \frisk{\clip{f}_{D_1, \lb, \g}}-\orisk
  &\leq \frac{c_1}{2} \Bparenth{\lb \g^{-d}+\g^{2\a}+\frac{(\log \lb^{-1})^{d+1}}{\g^{d} m}+\frac{\varrho}{m}}\\
  &\leq c_1\, \Bparenth{\lb \g^{-d}+\g^{2\a}+\frac{(\log \lb^{-1})^{d+1}}{\g^{d} n}+\frac{\varrho}{n}}\,,
 \end{align*}
with probability $P^m$ not less than $1-3 |\Lambda_n \times \Gamma_n |\, e^{-\varrho}$.
Now define $n-m \geq \frac{n}{2}-1 \geq \frac{n}{4}$ 
and $\varrho_n:= \varrho + \ln (1+|\Lambda_n \times \Gamma_n|)$, then by using \cite[Theorem 7.2]{steinwart2008support}
and Lemma \ref{Lemma-TV-SVM-RHS}, we obtain
\begin{align*}
&\frisk{\clip{f}_{D_1, \lb_{D_2}, \g_{D_2}}}-\orisk\\ 
&\leq 6  \infimum{(\lb, \g)\in \Lb_n,\G_n} \Bparenth{\frisk{\clip{f}_{D_1,\lb,\g}}-\orisk}+ 512 M^2 c_{\tau}^{-1} \frac{\varrho_n}{n-m}\\
& \leq 6c_1 \infimum{(\lb, \g)\in \Lb_n,\G_n} \Bparenth{\lb \g^{-d}+\g^{2\a}+\frac{(\log \lb^{-1})^{d+1}}{\g^{d} n}+\frac{\varrho}{n}}
+ 2048 M^2 c_{\tau}^{-1} \frac{\varrho_n}{n}\\
& \leq 6c_1 \Bparenth{c(\log n)^{d+1}\bparenth{n^{-\frac{2\a}{2\a+d}}+\delta_n^{2\a}}}+ 2048 M^2 c_{\tau}^{-1} \frac{\varrho_n}{n}\\
&\leq \varrho M^2 (\log n)^{d+1}(6c_1 c+6cc_1 \d_n^{2\a}+ 6c_1+2048  c_{\tau}^{-1} \varrho_n) n^{-\frac{2\a}{2\a+d}}\\
&\leq c_2 M^2 \varrho (\log n)^{d+1}n^{-\frac{2\a}{2\a+d}}\,,
\end{align*}
with probability $P^n$ not less than $1-3(1+\abs{\Lb_n\times \G_n})e^{-\varrho}$.
\end{proofof}

\begin{proofof}{Theorem \ref{Theorem-unbounded noise learning rates-newEntropy}}
By (\ref{Unbounded noise-entropy2}), we obtain
\begin{align*}
 P^n \Big(\Big\{D\in (X\times Y)^n:\underset{i\in\{1,\ldots,n\}}{\max}\{|y_i|\} \leq c \varrho^l\Big\}\Big) 
 &\geq  1-\sum_{i=1}^{n} P(|\epsilon_{y_i}|\geq c \varrho^l)\\
 & \geq 1-n e^{-\varrho}\\
 &= 1-e^{-(\varrho -\ln n)}\,.
\end{align*}
This implies that
\begin{equation*}
 P^n \Big(\Big\{D\in (X\times Y)^n:\underset{i\in\{1,\ldots,n\}}{\max}\{|y_i|\} \leq c(\hat{\varrho}+\ln n)^l\Big\}\Big) 
 \geq
 1-e^{-\hat{\varrho}}\,.
\end{equation*}
This leads us to conclude with probability $P^n$ not less than $1-e^{-\hat{\varrho}}$
that the SVM for ALS loss with belatedly clipped decision function 
at $M_n$ is actually a clipped regularized empirical risk minimization (CR-ERM) in the sense of 
\cite[Definition 7.18]{steinwart2008support}. 
 Consequently, \cite[Theorem 7.20]{steinwart2008support} holds
for $\hat {Y}:=\{-M_n,M_n\}$  modulo a set of probability $P^n$ not less than 
$1-e^{-\hat{\varrho}}$.
From Theorem \ref{Theorem-final_orale_inequlity-newEntropy}, we then obtain
\begin{equation*}
 \lb \snorm{f_{D, \lb, \g}}_{H_{\g}}^2 
 + \frisk{\clip{f}_{D, \lb, \g}}-\orisk \leq
  C M_n^{2}\,(\log \lb^ {-1})^{d+1}\Big(\lb \g^{-d}+ \g^{2\alpha}+n^{-1}\g^{-d}+n^{-1}\bar{\varrho} \Big)\,.
\end{equation*}
with probability $P^n$ not less than $1-e^{-\bar{\varrho}}-e^{-\hat{\varrho}}$. As in the proof of  Corollary 
(\ref{Corollary-lr_for_entropy_2}) and by using the inequality $(a+b)^c \leq (2ab)^c$, for $a,b\geq 1$ and $c > 0$, we finally obtain
\begin{align*}
 \lb \snorm{f_{D, \lb, \g}}_{H_{\g}}^2  + \frisk{\clip{f}_{D, \lb, \g}}-\orisk 
 &\leq
  C \bar{\varrho} M_n^{2} (\log n)^{d+1} n^{-\frac{2\alpha}{2\alpha+d}}\\
  &= C \bar{\varrho} \autoparenth{2c(\hat{\varrho}+\log n)^l}^2 (\log n)^{d+1} n^{-\frac{2\alpha}{2 \alpha+d}}\\
  &\leq  C \bar{\varrho}\,4c^2 \autoparenth{2\hat{\varrho}\,\log n}^{2l} (\log n)^{d+1} n^{-\frac{2\alpha}{2 \alpha+d}}\\
  & \leq \hat{C} \bar{\varrho}\hat{\varrho}^{2l} (\log n)^{2l+d+1} n^{-\frac{2\alpha}{2 \alpha+d}}\,,
\end{align*}
for all $n \geq 3$ with probability $P^n$ not less than 
	$1-e^{-\bar{\varrho}}-e^{-\hat{\varrho}}$. Choosing $\bar{\varrho}=\hat{\varrho}$ leads to the assertion.
\end{proofof}

\small{
\bibliographystyle{plain}
\bibliography{ms}

\begin{thebibliography}{10}

\bibitem{abdous1995relating}
B.~Abdous and B.~Remillard.
\newblock Relating quantiles and expectiles under weighted-symmetry.
\newblock {\em Ann. Inst. Statist. Math.}, 47:371--384, 1995.
\newblock \url{http://dx.doi.org/10.1007/bf00773468}.

\bibitem{adams2003sobolev}
R.~A. Adams and J.~J.~F. Fournier.
\newblock {\em Sobolev Spaces}.
\newblock Academic Press, New York, 2nd edition, 2003.
\newblock \url{https://doi.org/10.1016/s0079-8169(03)x8001-0}.

\bibitem{aragon2005conditional}
Y.~Aragon, S.~Casanova, R.~Chambers, and E.~Leconte.
\newblock Conditional ordering using nonparametric expectiles.
\newblock {\em J. Off. Stat.}, 21:617--633, 2005.
\newblock \url{http://www.jos.nu/Articles/abstract.asp?article=214617}.

\bibitem{aronszajn1950theory}
N.~Aronszajn.
\newblock Theory of reproducing kernels.
\newblock {\em Trans. Amer. Math. Soc.}, 68:337--404, 1950.
\newblock \url{}.

\bibitem{bauer2007regularization}
F.~Bauer, S.~Pereverzev, and L.~Rosasco.
\newblock On regularization algorithms in learning theory.
\newblock {\em J. complexity}, 23:52--72, 2007.
\newblock \url{https://doi.org/10.1016/j.jco.2006.07.001}.

\bibitem{BeKlMuGi14a}
F.~Bellini, B.~Klar, A.~M{\"u}ller, and R.~E. Gianin.
\newblock Generalized quantiles as risk measures.
\newblock {\em Insurance Math. Econom.}, 54:41--48, 2014.
\newblock \url{ http://dx.doi.org/10.1016/j.insmatheco.2013.10.015 }.

\bibitem{blanchard2008statistical}
G.~Blanchard, O.~Bousquet, and P.~Massart.
\newblock Statistical performance of support vector machines.
\newblock {\em Ann. Statist.}, pages 489--531, 2008.
\newblock \url{ https://doi.org/10.1214/009053607000000839}.

\bibitem{breckling1988m}
J.~Breckling and R.~Chambers.
\newblock M-quantiles.
\newblock {\em Biometrika}, 75:761--771, 1988.
\newblock \url{http://dx.doi.org/10.2307/2336317 }.

\bibitem{caponnetto2007optimal}
A.~Caponnetto and E.~De~Vito.
\newblock Optimal rates for the regularized least-squares algorithm.
\newblock {\em Found. Comput. Math.}, 7:331--368, 2007.
\newblock \url{ https://doi.org/10.1007/s10208-006-0196-8}.

\bibitem{chen2004support}
D.R. Chen, Q.~Wu, Y.~Ying, and D.X. Zhou.
\newblock Support vector machine soft margin classifiers: error analysis.
\newblock {\em J. Mach. Learn. Res.}, 5:1143--1175, 2004.
\newblock \url{}.

\bibitem{christmann2007svms}
A.~Christmann and I.~Steinwart.
\newblock How {SVM}s can estimate quantiles and the median.
\newblock In {\em Advances in neural information processing systems}, pages
  305--312, 2007.

\bibitem{cucker2002mathematical}
F.~Cucker and S.~Smale.
\newblock On the mathematical foundations of learning.
\newblock {\em Bull. Amer. Math. Soc.}, 39:1--49, 2002.
\newblock \url{https://doi.org/10.1090/s0273-0979-01-00923-5}.

\bibitem{de2005model}
E.~De~Vito, A.~Caponnetto, and L.~Rosasco.
\newblock Model selection for regularized least-squares algorithm in learning
  theory.
\newblock {\em Foun. Comput. Math.}, 5:59--85, 2005.
\newblock \url{https://doi.org/10.1007/s10208-004-0134-1}.

\bibitem{devore1998nonlinear}
R.~A. DeVore.
\newblock Nonlinear approximation.
\newblock {\em Acta numerica}, 7:51--150, 1998.
\newblock \url{}.

\bibitem{devore1988interpolation}
R.~A. DeVore and V.~A. Popov.
\newblock Interpolation of {B}esov spaces.
\newblock {\em Trans. Amer. Math. Soc.}, 305:397--414, 1988.
\newblock \url{}.

\bibitem{devore1993besov}
R.~A. DeVore and R.~C. Sharpley.
\newblock {B}esov spaces on domains in $\mathbb{R}^d$.
\newblock {\em Trans. Amer. Math. Soc.}, 335:843--864, 1993.
\newblock \url{}.

\bibitem{eberts2013optimal}
M.~Eberts and I.~Steinwart.
\newblock Optimal regression rates for {SVM}s using {G}aussian kernels.
\newblock {\em Electron. J. Stat.}, 7:1--42, 2013.
\newblock \url{http://dx.doi.org/10.1214/12-ejs760}.

\bibitem{edmunds2008function}
D.~E. Edmunds and H.~Triebel.
\newblock {\em Function Spaces, Entropy Numbers, Differential Operators}.
\newblock Cambridge University Press, Cambridge, 2008.
\newblock \url{ https://doi.org/10.1017/cbo9780511662201}.

\bibitem{efron1991regression}
B.~Efron.
\newblock Regression percentiles using asymmetric squared error loss.
\newblock {\em Statist. Sci.}, 1:93--125, 1991.
\newblock \url{}.

\bibitem{farooq2015svm}
M.~Farooq and I.~Steinwart.
\newblock An {SVM}-like approach for expectile regression.
\newblock {\em Comput. Stat. Data Anal.}, 109:159--181, 2017.
\newblock \url{https://doi.org/10.1016/j.csda.2016.11.010}.

\bibitem{gneiting2011making}
T.~Gneiting.
\newblock Making and evaluating point forecasts.
\newblock {\em J. Am. Stat. Assoc.}, 106:746--762, 2011.
\newblock \url{http://dx.doi.org/10.1198/jasa.2011.r10138}.

\bibitem{guler2014mincer}
K.~Guler, P.~T. Ng, and Z.~Xiao.
\newblock Mincer-{Z}arnovitz quantile and expectile regressions for forecast
  evaluations under asymmetric loss functions.
\newblock {\em Northern Arizona University, The WA Franke College of Business.
  Working Paper Series 14-01}, 2014.
\newblock \url{}.

\bibitem{hamidi2014dynamic}
B.~Hamidi, B.~Maillet, and J-L. Prigent.
\newblock A dynamic autoregressive expectile for time-invariant portfolio
  protection strategies.
\newblock {\em J. Econom. Dynam. Control}, 46:1--29, 2014.
\newblock \url{http://dx.doi.org/10.1016/j.jedc.2014.05.005}.

\bibitem{hush2006qp}
D.~Hush, P.~Kelly, C.~Scovel, and I.~Steinwart.
\newblock {QP} algorithms with guaranteed accuracy and run time for support
  vector machines.
\newblock {\em J. Mach. Learn. Res.}, 7:733--769, 2006.
\newblock \url{}.

\bibitem{kim2016nonlinear}
M.~Kim and S.~Lee.
\newblock Nonlinear expectile regression with application to value-at-risk and
  expected shortfall estimation.
\newblock {\em Comput. Stat. Data Anal.}, 94:1--19, 2016.
\newblock \url{https://doi.org/10.1016/j.csda.2015.07.011}.

\bibitem{koenker1978regression}
R.~Koenker and G.~Bassett~Jr.
\newblock Regression quantiles.
\newblock {\em Econometrica}, 46:33--50, 1978.
\newblock \url{http://dx.doi.org/10.2307/1913643}.

\bibitem{meister2016optimal}
M.~Meister and I.~Steinwart.
\newblock Optimal learning rates for localized {SVM}s.
\newblock {\em J. Mach. Learn. Res.}, 17:1--44, 2016.
\newblock \url{}.

\bibitem{mendelson2010regularization}
S.~Mendelson and J.~Neeman.
\newblock Regularization in kernel learning.
\newblock {\em Ann. Statist.}, 38:526--565, 2010.
\newblock \url{https://doi.org/10.1214/09-aos728}.

\bibitem{newey1987asymmetric}
W.~K. Newey and J.~L. Powell.
\newblock Asymmetric least squares estimation and testing.
\newblock {\em Econometrica}, 55:819--847, 1987.
\newblock \url{http://dx.doi.org/10.2307/1911031}.

\bibitem{nikol2012approximation}
S.~M. Nikol'skii.
\newblock {\em Approximation of Functions of Several Variables and Imbedding
  Theorems}, volume 205.
\newblock Springer Science \& Business Media, 2012.
\newblock \url{ https://doi.org/10.1007/978-3-642-65711-5}.

\bibitem{schnabel2009analysis}
S.~Schnabel and P.~Eilers.
\newblock An analysis of life expectancy and economic production using
  expectile frontier zones.
\newblock {\em Demographic Res.}, 21:109--134, 2009.
\newblock \url{http://dx.doi.org/10.4054/demres.2009.21.5}.

\bibitem{sobotka2012geoadditive}
F.~Sobotka and T.~Kneib.
\newblock Geoadditive expectile regression.
\newblock {\em Comput. Statist. Data Anal.}, 56:755--767, 2012.
\newblock \url{http://dx.doi.org/10.1016/j.csda.2010.11.015}.

\bibitem{sobotka2013estimating}
F.~Sobotka, R.~Radice, G.~Marra, and T.~Kneib.
\newblock Estimating the relationship between women's education and fertility
  in {B}otswana by using an instrumental variable approach to semiparametric
  expectile regression.
\newblock {\em J. Roy. Stat. Soc. C- App.}, 62:25--45, 2013.
\newblock \url{http://dx.doi.org/10.1111/j.1467-9876.2012.01050.x}.

\bibitem{stahlschmidt2014expectile}
S.~Stahlschmidt, M.~Eckardt, and W.~K. H{\"a}rdle.
\newblock Expectile treatment effects: {A}n efficient alternative to compute
  the distribution of treatment effects.
\newblock Technical report, Sonderforschungsbereich 649, Humboldt University,
  Berlin, Germany, 2014.

\bibitem{steinwart2007compare}
I.~Steinwart.
\newblock How to compare different loss functions and their risks.
\newblock {\em Constr. Approx.}, 26:225--287, 2007.
\newblock \url{https://doi.org/10.1007/s00365-006-0662-3}.

\bibitem{steinwart2009oracle}
I.~Steinwart.
\newblock Oracle inequalities for support vector machines that are based on
  random entropy numbers.
\newblock {\em J. Complexity}, 25:437--454, 2009.
\newblock \url{https://doi.org/10.1016/j.jco.2009.06.002}.

\bibitem{steinwart2008support}
I.~Steinwart and A.~Christmann.
\newblock {\em Support Vector Machines}.
\newblock Springer, New York, 2008.
\newblock \url{ http://dx.doi.org/10.1007/978-0-387-77242-4}.

\bibitem{steinwart2011estimating}
I.~Steinwart and A.~Christmann.
\newblock Estimating conditional quantiles with the help of the pinball loss.
\newblock {\em Bernoulli}, 17:211--225, 2011.
\newblock \url{http://dx.doi.org/10.3150/10-bej267}.

\bibitem{steinwart2006oracle}
I.~Steinwart, D.~Hush, and C.~Scovel.
\newblock An oracle inequality for clipped regularized risk minimizers.
\newblock In {\em Advances in neural information processing systems}, pages
  1321--1328, 2006.

\bibitem{steinwart2011training}
I.~Steinwart, D.~Hush, and C.~Scovel.
\newblock Training {SVM}s without offset.
\newblock {\em J. Mach. Learn. Res.}, 12:141--202, 2011.
\newblock \url{}.

\bibitem{steinwart2009optimal}
I.~Steinwart, D.~R. Hush, and C.~Scovel.
\newblock Optimal rates for regularized least squares regression.
\newblock In {\em 22nd Annual Conference on Learning Theory}, pages 79--93,
  2009.

\bibitem{StPaWiZh14a}
I.~Steinwart, C.~Pasin, R.~Williamson, and S.~Zhang.
\newblock Elicitation and identification of properties.
\newblock In M.~F. Balcan and C.~Szepesvari, editors, {\em JMLR Workshop and
  Conference Proceedings Volume 35: Proceedings of the 27th Conference on
  Learning Theory 2014}, pages 482--526, 2014.

\bibitem{Steinwart14b}
I.~Steinwart and P.~Thomann.
\newblock {liquidSVM}: A fast and versatile {SVM} package.
\newblock Technical report, Fakult{\"a}t f{\"u}r Mathematik und Physik,
  Universit{\"at} Stuttgart, 2017.
\newblock \url{https://arxiv.org/abs/1702.06899}.

\bibitem{tacchetti2013gurls}
A.~Tacchetti, P.~K. Mallapragada, M.~Santoro, and L.~Rosasco.
\newblock {GURLS}: a least squares library for supervised learning.
\newblock {\em J. Mach. Learn. Res.}, 14:3201--3205, 2013.
\newblock \url{}.

\bibitem{tartar2007introduction}
L.~Tartar.
\newblock {\em An Introduction to Sobolev Spaces and Interpolation Spaces},
  volume~3.
\newblock Springer Science \& Business Media, 2007.
\newblock \url{https://doi.org/10.1007/978-3-540-71483-5}.

\bibitem{taylor2008estimating}
J.~W. Taylor.
\newblock Estimating value at risk and expected shortfall using expectiles.
\newblock {\em J. Financ. Econ.}, 6:231--252, 2008.
\newblock \url{ http://dx.doi.org/10.1093/jjfinec/nbn001}.

\bibitem{van2009adaptive}
A.~W. van~der Vaart and J.~H. van Zanten.
\newblock Adaptive {B}ayesian estimation using a {G}aussian random field with
  inverse {G}amma bandwidth.
\newblock {\em Ann. Statist.}, 37:2655--2675, 2009.
\newblock \url{ https://doi.org/10.1214/08-aos678}.

\bibitem{wang2011measuring}
Y.~Wang, S.~Wang, and K.~K. Lai.
\newblock Measuring financial risk with generalized asymmetric least squares
  regression.
\newblock {\em Appl. Soft Comput.}, 11(8):5793--5800, 2011.
\newblock \url{ http://dx.doi.org/10.1016/j.asoc.2011.02.018}.

\bibitem{wu2006learning}
Q.~Wu, Y.~Ying, and D-X. Zhou.
\newblock Learning rates of least-square regularized regression.
\newblock {\em Found. Comput. Math.}, 6:171--192, 2006.
\newblock \url{https://doi.org/10.1007/s10208-004-0155-9}.

\bibitem{xu2016nonparametric}
Q.~Xu, X.~Liu, C.~Jiang, and K.~Yu.
\newblock Nonparametric conditional autoregressive expectile model via neural
  network with applications to estimating financial risk.
\newblock {\em Appl. Stoch. Model Bus.}, 32:882--908, 2016.
\newblock \url{https://doi.org/10.1002/asmb.2212}.

\bibitem{yang2014nonparametric}
Y.~Yang and H.~Zou.
\newblock Nonparametric multiple expectile regression via {ER}-{B}oost.
\newblock {\em J. Stat. Comput. Simulation}, 85:1442--1458, 2015.
\newblock \url{http://dx.doi.org/10.1080/00949655.2013.876024}.

\bibitem{yao1996asymmetric}
Q.~Yao and H.~Tong.
\newblock Asymmetric least squares regression estimation: a nonparametric
  approach.
\newblock {\em J. Nonparametr. Statist.}, 6:273--292, 1996.
\newblock \url{http://dx.doi.org/10.1080/10485259608832675}.

\bibitem{ziegel2014coherence}
J.~F. Ziegel.
\newblock Coherence and elicitability.
\newblock {\em Math. Financ.}, 26:901--918, 2016.
\newblock \url{https://doi.org/10.1111/mafi.12080}.

\end{thebibliography}
}

\end{document}